\documentclass[10pt,twocolumn,letterpaper]{article}

\usepackage[pagenumbers]{wacv} 

\usepackage{graphicx}
\usepackage{amsmath}
\usepackage{amssymb}
\usepackage{booktabs}
\usepackage{multirow}
\usepackage{dcolumn}
\usepackage{longtable}
\usepackage{tcolorbox}
\usepackage{enumitem}
\usepackage{minitoc}

\usepackage[pagebackref,breaklinks,colorlinks]{hyperref}
\newcommand{\NA}{---}

\usepackage[capitalize]{cleveref}
\crefname{section}{Sec.}{Secs.}
\Crefname{section}{Section}{Sections}
\Crefname{table}{Table}{Tables}
\crefname{table}{Tab.}{Tabs.}

\def\wacvPaperID{2289} 
\def\confName{WACV}
\def\confYear{2025}

\begin{document}

\title{Learning Multiple Object States from Actions via Large Language Models}

\author{Masatoshi Tateno$^{1,2}$
\qquad
Takuma Yagi$^{2}$
\qquad
Ryosuke Furuta$^{1}$
\qquad
Yoichi Sato$^{1}$
\\
$^{1}$ The University of Tokyo \\ $^{2}$ National Institute of Advanced Industrial Science and Technology (AIST)\\
{\tt\small \{masatate, furuta, ysato\}@iis.u-tokyo.ac.jp, takuma.yagi@aist.go.jp}\\
\small{\url{https://masatate.github.io/ObjStatefromAction.github.io/}}
}
\maketitle

\doparttoc 
\faketableofcontents 


\begin{abstract}
Recognizing the states of objects in a video is crucial in understanding the scene beyond actions and objects. For instance, an egg can be {\it raw}, {\it cracked}, and {\it whisked} while cooking an omelet, and these states can coexist simultaneously (an egg can be both \textit{raw} and \textit{whisked}). However, most existing research assumes single object state change (\eg uncracked $\rightarrow$ cracked), overlooking the coexisting nature of multiple object states and the influence of past states on the current state. We formulate object state recognition as a multi-label classification task that explicitly handles multiple states. We then propose to learn multiple object states from narrated videos by leveraging LLMs to generate pseudo-labels from the transcribed narrations, capturing the influence of past states. The challenge is that narrations mostly describe human actions in the video but rarely explain object states. Therefore, we use LLM's knowledge of the relationship between actions and states to derive the missing object states. We further accumulate the derived object states to consider the past state contexts to infer current object state pseudo-labels. We newly collect Multiple Object States Transition (MOST) dataset, which includes manual multi-label annotation for evaluation purposes, covering 60 object states across six object categories. Experimental results show that our model trained on LLM-generated pseudo-labels significantly outperforms strong vision-language models, demonstrating the effectiveness of our pseudo-labeling framework that considers past context via LLMs.
\end{abstract}
\section{Introduction}
\label{sec:intro}

\begin{figure}[tb]
    \centering
    \includegraphics[width=1\linewidth]{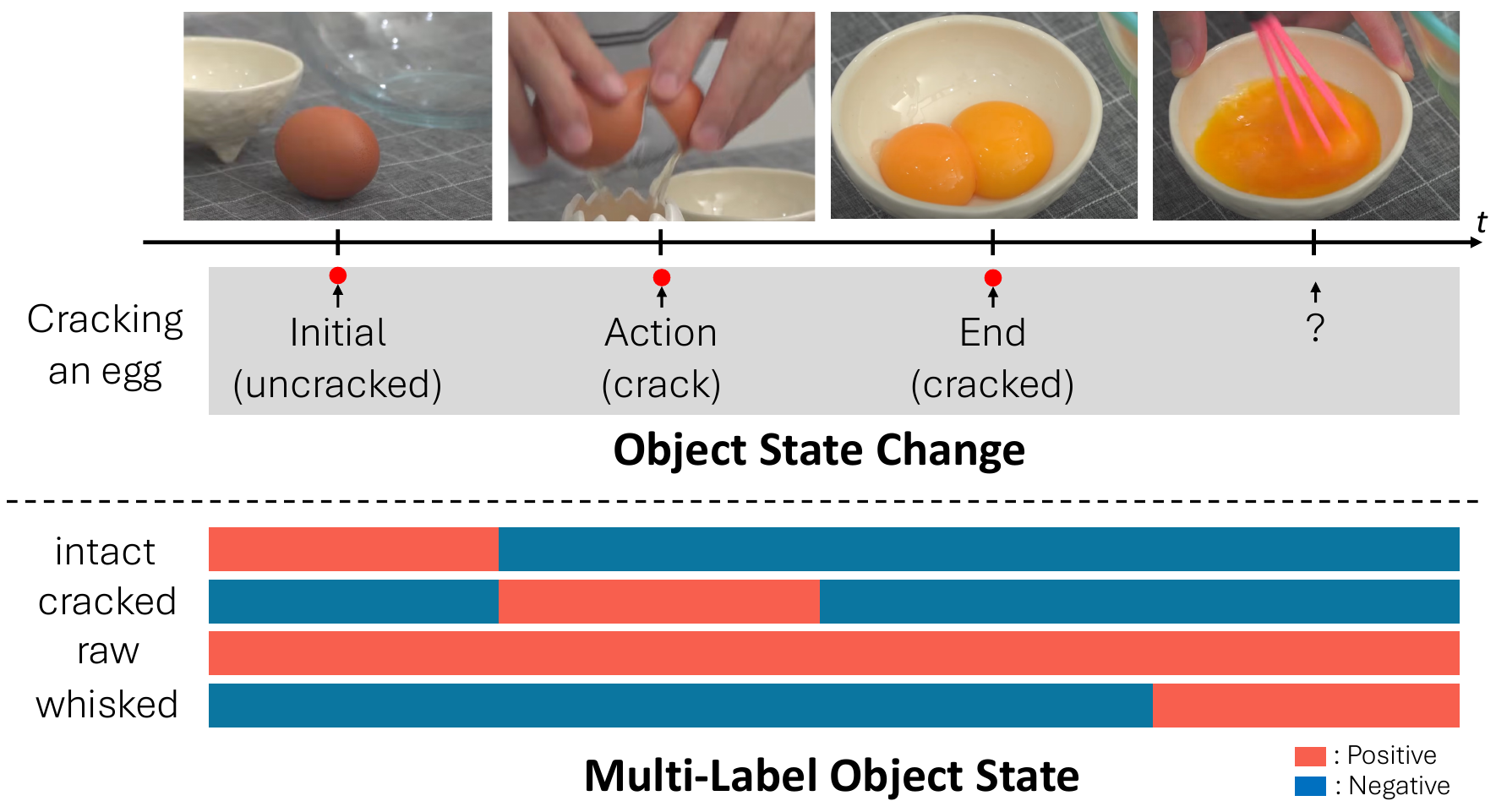}
    \caption{We formulate object state recognition as a multi-label frame-wise classification task (bottom). Compared to formulation focusing on state change in previous works (top), our formulation explicitly considers coexisting multiple object states.}
    \label{fig:teaser}
\end{figure}

What do we need to understand in a scene of cooking an omelet?
Actions and objects involved in the scene, like ``crack eggs into a bowl'' are some key elements, but not all.
As the scene progresses, objects undergo multiple state changes triggered by human actions.
For example, when one cracks an egg and then whisks it, the {\it intact} egg becomes {\it cracked}, and then becomes {\it whisked}, while it remains {\it raw} during these actions (Figure \ref{fig:teaser}). Recognizing the existence of these multiple object states ({\it intact}, {\it cracked}, {\it whisked}, {\it raw}) is crucial for accurately understanding the progression of procedural tasks across various domains, including cooking~\cite{nishimura2021state}, assembly~\cite{damen2016you,sener2022assembly101}, and robotics~\cite{koppula2013anticipating}.


However, previous research primarily focuses on single object state change in a video \cite{joint_discovery ,look_for_the_change, multi_task, xue2023learning}, assuming only one state change (\eg uncracked $\rightarrow$ cracked) always occurs in a video. 
The task they tackle is temporal localization of state transitions (\textit{initial state}~/~\textit{action}~/~\textit{end state}) (Figure \ref{fig:teaser} top) rather than focusing on the existence of the object states themselves (\eg~{\it cracked}, {\it whisked}, {\it raw}) (Figure \ref{fig:teaser} bottom).
Although the single state change assumption allows them to use ordering constraints (\textit{initial} $\rightarrow$ \textit{action} $\rightarrow$ \textit{end}) that help tackle the localization task, it is inadequate for understanding human activities involving multiple state changes or when an object state remains unchanged. 
Moreover, previous works overlook how the previous state of an object influence the outcome of an action (\eg, heating an \textit{uncracked egg} and heating a \textit{whisked egg} yields different outcomes).



We reformulate object state recognition as a \textbf{multi-label frame-wise classification task}, accounting for videos that may include several state changes of a particular class of object.
Given any object state, our goal is to train a recognition model that predicts whether each object state exists at each moment of the video.
The multi-label setting requires capturing coexisting multiple object states by disentangling complex state transitions in a scene while considering the influence of past states.

Due to its complexity, it is labor-intensive to manually collect large-scale multi-label annotation for model training. To bypass the annotation effort, learning from transcribed narrations in instructional videos is intriguing. Narrations provide valuable information about what is happening in videos and can serve as a scalable resource for training models. However, these narrations often focus on describing actions, lacking detailed explanations of object states.

Therefore, we propose to leverage the LLM’s knowledge of the relationship between actions and states to derive the missing object state information, while representing past states as state descriptions to consider their influence.
Our observation is that action
information and past object states implies the existence of current object states. State descriptions, which concisely describe object states after each action, provide a more general representation than state labels, allowing consideration of past states.
Our pseudo-labeling framework is three steps: (i) first extracting actions mentioned in the narrations, (ii) using the past state and the current action to describe the object’s new state, and (iii) accumulating these past state descriptions to infer object state labels. We also correct the temporal misalignment between video frames, inferred actions, and object states—caused by the misalignment between video and narration—using off-the-shelf VLMs.

We then use LLM-generated pseudo-labels to train multi-label temporal segmentation models. However, some pseudo-labels are missing due to ambiguity in inferring states from actions, which we find yields suboptimal results. Therefore, we adopt self-training to fill in the gap. Specifically, we find that using an ensemble of two models with different context lengths as teacher models effectively produces better soft labels for training student models.

We collect a new Multiple Object States Transition (MOST) dataset to evaluate the temporal segmentation of multi-label object states. The dataset includes multi-label object state annotation for each of six object categories (apple, egg, flour, shirt, tire, and wire), covering 60 distinct states in total.
The MOST dataset is designed to evaluate the existence of object states, different from the previous datasets~\cite{look_for_the_change,grauman2022ego4d,xue2023learning} that aim to find the timing of object state change.
Experimental evaluation shows that our model trained on LLM-generated pseudo-labels achieves a significant improvement of over 29\% in mAP against strong vision-language models~\cite{clip,wang2022internvideo}, demonstrating the superiority in capturing the multiple object states, which is enabled by our context-aware framework that derives accurate object states from narrations via LLMs.

\section{Related Work}
\label{sec:related_work}


\subsection{Learning Object States from Videos}
Learning and recognizing object states have been studied using videos since they capture the dynamics of object state changes better than images.
Tasks such as object state classification~\cite{liu2017jointly,fire2017inferring}, discovery of internal states~\cite{damen2016you,joint_discovery}, action-state modeling~\cite{fathi2013modeling,zhuo2019explainable}, and point-of-no-return temporal localization~\cite{grauman2022ego4d} have been studied.
While early works relied on small-scale video collection and annotation, learning from abundant Internet videos with video captions has become a trend due to its scalability~\cite{bertasius2020cobe,look_for_the_change,multi_task}.
Sou\v{c}ek \etal~\cite{look_for_the_change} propose an unsupervised method to learn the object state change dynamics from Internet videos using causal ordering constraints, further extended to multi-task learning~\cite{multi_task}.
While ordering constraints provide natural supervision for object state recognition, it is based on assumptions that (i) the state-changing action always exists and (ii) the object state change occurs only once in a video—conditions that may not hold in practical scenarios. On the contrary, our work utilize LLMs to  extract object state information from narrations in a scalable manner without relying on heuristics.
Concurrently to our work, Xue \etal~\cite{xue2023learning} also uses LLMs to mine the possible object with specific state transitions from video narrations in an open-world manner.
The key difference is that our work focuses on the multi-label nature of object states, which requires an understanding of how past states influence current ones. We propose a framework to fully exploit the narrations to capture this influence.


\subsection{Learning from Video Narrations}
The transcribed narrations of Internet videos can be used as a scalable resource for video learning tasks~\cite{sener2015unsupervised,alayrac2016unsupervised,zhou2018towards,sun2019videobert,howto100m,zhukov2019cross,miech2020end,shen2021learning,lin2022learning,xue2022advancing,dvornik2023stepformer,zhou2023procedure,zhong2023learning}.
However, these ASR narrations are found to be suboptimal due to irrelevant or noisy descriptions (\eg, introduction, random chat, advertisement) and erroneous temporal boundaries.
To mitigate the above issue, techniques such as multiple instance learning~\cite{miech2020end}, distant supervision via wikiHow~\cite{lin2022learning}, use of knowledge graph~\cite{zhou2023procedure}, learning the alignability of narrations~\cite{han2022temporal}, sequence-to-sequence alignment~\cite{dvornik2023stepformer}, and temporal modeling~\cite{zhong2023learning}, have been proposed.
The emphasis of these methods has been on verbs and nouns in narration. Learning of adjectives, including states, has only been addressed to a limited extent.
Instead of removing or replacing irrelevant information from narrations as in previous work, we attempt to extract object state information from narrations, using the internal knowledge of LLMs.

\subsection{LLMs for Visual Learning}
LLMs trained by large amounts of text corpus (\eg, \cite{gpt-3}) have been demonstrating remarkable ability in various text reasoning tasks.
Recently, several attempts to leverage LLMs to visual learning are made in tasks as image captioning~\cite{chen2022visualgpt,li2023blip}, action recognition~\cite{lin2023match}, video understanding~\cite{wang2022language,zeng2022socratic,wang2023internvid}, and active object grounding~\cite{wu2023localizing}
In video-text tasks, the high flexibility of LLMs allows us to extract essential information from noisy narratives~\cite{wang2023internvid} and to expand text labels by summarization and rephrasing~\cite{zhao2023learning}. 
Lin \etal~\cite{lin2023match} propose an unsupervised approach using off-the-shelf VLMs and LLMs to construct a text bag corresponding to unlabeled videos, which is optimized by a multiple-instance learning objective.
LAVILA~\cite{zhao2023learning} uses LLMs to densely describe the events in long-form videos.
Alper \etal~\cite{alper2023learning} propose a similar scheme of producing free-form text labels through LLMs for human-human interaction prediction problems.
Instead of using LLMs for simple rephrasing or captioning, we propose to use LLMs as a bridge between two different concepts---actions and object states.
We show that information not always mentioned often, \ie  object states, can be efficiently recovered from label-abundant concepts such as actions.
\section{Problem Setting}
\label{sec:problem_setting}
Our goal is to predict the existence of multiple object states at each moment of a video.
We formulate the problem as a multi-label frame-wise classification task.

Formally, we assume a set of instructional videos $\mathcal{V}$, where each video $\mathbf{X} \in \mathcal{V}$ is accompanied by a set of video narrations describing the content of the video.
Given a video $\mathbf{X}=\{\mathbf{x}_1, \mathbf{x}_2, \dots, \mathbf{x}_T\}$ of length $T$, where $\mathbf{x}_i$ denotes the $i$-th video frame, the goal is to predict a sequence of binary vectors $\mathbf{Y} = \{\mathbf{y}_1, \mathbf{y}_2, \dots, \mathbf{y}_T\}$, where $\mathbf{y}_i = \{y_i^1, y_i^2, ..., y_i^K\}$ is a binary vector denoting the presence of $K$ object state labels at time $i$.

Since there is often a range of meanings for an object state, we assume pairs of a state name (\eg, \textit{sliced}) and its textual description composed of a few sentences (\eg, \textit{``the state of an apple that has been cut into thin or narrow pieces. ... applicable as long as it retains the shape of the pieces.''}).
Object state labels are NOT mutually exclusive and more than one object state may appear in a video frame simultaneously (\eg, a shirt may be {\it soaked} and {\it dyed} at the same time). If the object in focus does not appear on the screen, we assume all state labels are negative.

\section{Method}
\label{sec:method}
In this section, we propose a method to learn multiple object states. The challenge is the lack of training data. Manually annotating multiple object states requires a huge effort to simultaneously track different states appearing and disappearing, making it difficult to collect large-scale. To bypass the annotation effort, learning from the narrated instructional videos ~\cite{alayrac2016unsupervised,miech2020end,bertasius2020cobe} has been a promising way in various video recognition tasks.

However, in instructional videos, speakers talk about the most important events first, and often omit obvious or unimportant information~\cite{gordon2013reporting,misra2016seeing}.
They often do not tell much about the states of objects compared to what they are doing (actions) and what they are handling (objects).
Therefore, naively training object state recognition models from raw narrations may lead to suboptimal results.

To address the lack of object state information in the narrations, we introduce to leverage LLMs to generate object state pseudo-labels based on the narrations to train the recognition model. In the following sections, we first explain how to extract object state labels from narrations (Section \ref{subsec:label_extraction}) and temporally align the inferred pseudo-labels with video frames (Section \ref{subsec:aligning_object_states}). Then, we explain the training process using generated pseudo-labels in Section \ref{subsec:training}.

\subsection{Object State Label Extraction from Narrations via LLMs}
\label{subsec:label_extraction}

\begin{figure*}[tb]
    \centering
    \includegraphics[width=1\linewidth]{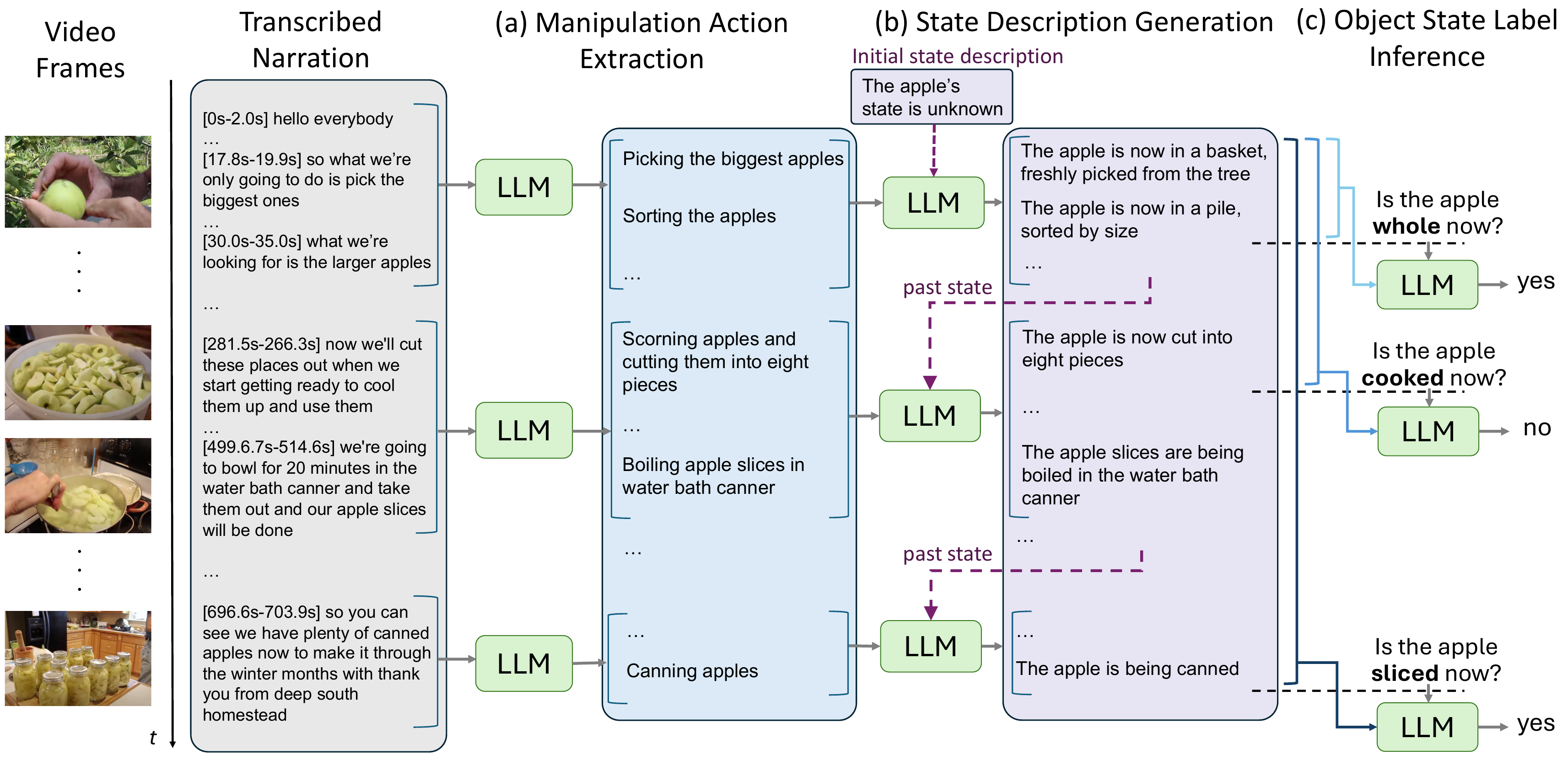}
    \caption{Three-stage framework of inferring presence of object state labels from video narrations. (a) Manipulation action extraction from narrations. (b) Object state description generation from actions. (c) Context-aware object state label inference.}
    \label{fig:llm_part}
\end{figure*}



To address the lack of object state information in the narrations, We introduce a framework to generate object state pseudo-labels based on action information in the narrations using LLMs. The key observation is that the existence of each object state, which is not necessarily mentioned in the narrations, can be inferred from (1) the actor's current actions, and (2) the past state of the object. Thus, our framework consists of three stages as shown in Figure \ref{fig:llm_part}: (a) manipulation action extraction, (b) state description generation from actions, and (c) context-aware object state label inference.
\vspace{0.6em}

\noindent \textbf{Manipulation Action Extraction}
The goal of this stage is to extract the manipulation actions of the actor from ASR narrations.
Raw ASR transcriptions paired with instructional videos offer valuable information on the actor's actions.
However, raw narrations often contain transcription errors, over-segmentation, and irrelevant content unrelated to the visible actions, which may harm inferring object states.
To this end, we leverage LLMs to extract {\it manipulation actions} actually performed in the scene, from narrations. Specifically, a block of narrations (10 sentences per block) is fed to LLMs, which output manipulation actions (Figure~\ref{fig:llm_part} (a)). The extracted actions will be used as key points to consider object states in the subsequent step.

Note that manipulation actions do not correspond to original narrations one by one.
To obtain the time interval for each extracted action, we also prompt LLMs to refer back to mark the narrations that best match what the action says, and take the interval of it as the interval of the manipulation action.
\vspace{0.6em}

\noindent \textbf{Object State Description Generation}
Next, we sequentially generate a set of {\it state description} from manipulation actions, which is used as intermediate information before generating the multi-label object state pseudo-labels (Figure~\ref{fig:llm_part} (b)).
State description is a short sentence including the target object name and a description concisely describing their state.
The motivation of state description is to keep track of the object's state at that time while detecting whether the object has undergone any kind of state change.
The accumulation of these descriptions will serve as the rich past context of object states to consider its influence in the subsequent label inference step.

In this stage, we assume a set of action blocks, each including 10 manipulation actions, and the initial state description is set as ``The state of $<$object name$>$ is unknown.''.
At each block, we pass the set of current manipulation actions and the last state description to LLMs and generate a sequence of state descriptions corresponding to each action.
By considering the past state description, LLMs can track what change has been applied to the object, and generate temporally consistent state descriptions across videos.
\vspace{0.6em}


\noindent \textbf{Context-aware Object State Label Inference}
Finally, we infer whether a specific target object state (\eg, whole, cooked, or sliced) exists at the timing of each state description. (Figure~\ref{fig:llm_part} (c))
For each state description, we feed the concatenation of all the state descriptions up to that timing to LLMs and prompt it to produce the existence (in either ``Yes'', ``No'' or ``ambiguous'') of each of the $K$ object states.

The accumulation of the state descriptions contains a rich implicature regarding the existence of each object state. It makes it possible to consider how the current object state is influenced by the past states. For example, if we want to determine whether the apple is sliced, we need to check whether it has been sliced before and make sure the shape of the slice has not been lost until that time. 
When inferring object state labels, we also input their textual description as described in Section \ref{sec:problem_setting} to LLMs to clarify the state's meaning.
We allow LLMs to answer ``ambiguous'' if the given list of state descriptions does not contain enough information.
The loss is not calculated for the object state labels determined as ambiguous during training.

\subsection{Interval Alignment}
\label{subsec:aligning_object_states}

After we obtain the list of object state labels with their corresponding manipulation actions, action time intervals, and state descriptions, we use VLM to align the object states with video frames.
The time intervals of object states are likely to be contained within the intervals of actions.
However, they are not correctly temporally aligned due to misalignment between (a) narrations and video frames (b) the existence of object states and actions. 
Therefore, we use VLMs to first align the video frame with the correct action, and then use VLMs again to determine if the corresponding object state description is true for the video frame. If a state description is not aligned with the video frame, the corresponding state labels will not be assigned, and we will not calculate the loss in the subsequent training stage. If we could assign a state description for the video frame, we assign the corresponding state labels to the video frame. See supplementary for the details.

\subsection{Self-Training from Pseudo-Labels}
\label{subsec:training}

After the interval alignment, we obtain object state pseudo-labels to train the recognition model. However, we found that learning directly from the pseudo-labels leads to suboptimal results since they are occasionally not assigned due to ambiguity or unalignability with the video frames. Thus, we apply a self-training scheme where teacher models learned from temporally sparse labels produce labels to train student models without missing any labels. 

Specifically, we adopt mean teacher learning~\cite{tarvainen2017mean}, which updates the target student model using the ensemble of outputs from teacher models initialized with original pseudo-labels. Both teacher and student models consist of two frame-wise classification models: \textbf{MLP} and \textbf{TCN}, each with different temporal windows. They share a frozen backbone from Internvideo~\cite{wang2022internvideo}, but differ in classifiers—Multi-Layer Perceptron for one, Temporal Convolutional Networks for the other. These two models play complementary roles in recognizing object states, and thereby, the ensemble of two soft labels better mitigates the missing label issue. The parameters of the teacher model are updated by copying those of the student model using Exponential Moving Averages. After the training, only the student {\bf TCN} model is used for inference. See supplementary for the details.


\section{Experiments}
\label{sec:experiments}

We examine our proposed approach in the newly collected Multiple Object States Transition (MOST) dataset. Additionally, we conduct supplementary evaluations on the existing ChangeIt dataset~\cite{look_for_the_change} to demonstrate that our framework also performs comparably in the state change localization task.

\subsection{Multiple Object States Transition Dataset (MOST)}
\label{subsec:dataset}

Since there were no publicly available datasets on the temporal segmentation of multi-label object states, we created a new evaluation dataset called Multiple Object States Transition (MOST) dataset.
We manually collected instructional videos from YouTube for six object categories: apple, egg, flour, shirt, tire, and wire, that take various types of object states.
We selected around 10 object states per object category to annotate, covering the popular states the object may take (see supplementary for full lists).
For the videos showing the target object, annotators marked the interval where each object state exists.
As a result, 61 videos with a total duration of 159.6 minutes were fully annotated.

\begin{figure}[tb]
  \centering
  \begin{subfigure}{1\linewidth}
    \centering
    \includegraphics[width=1.0\linewidth]{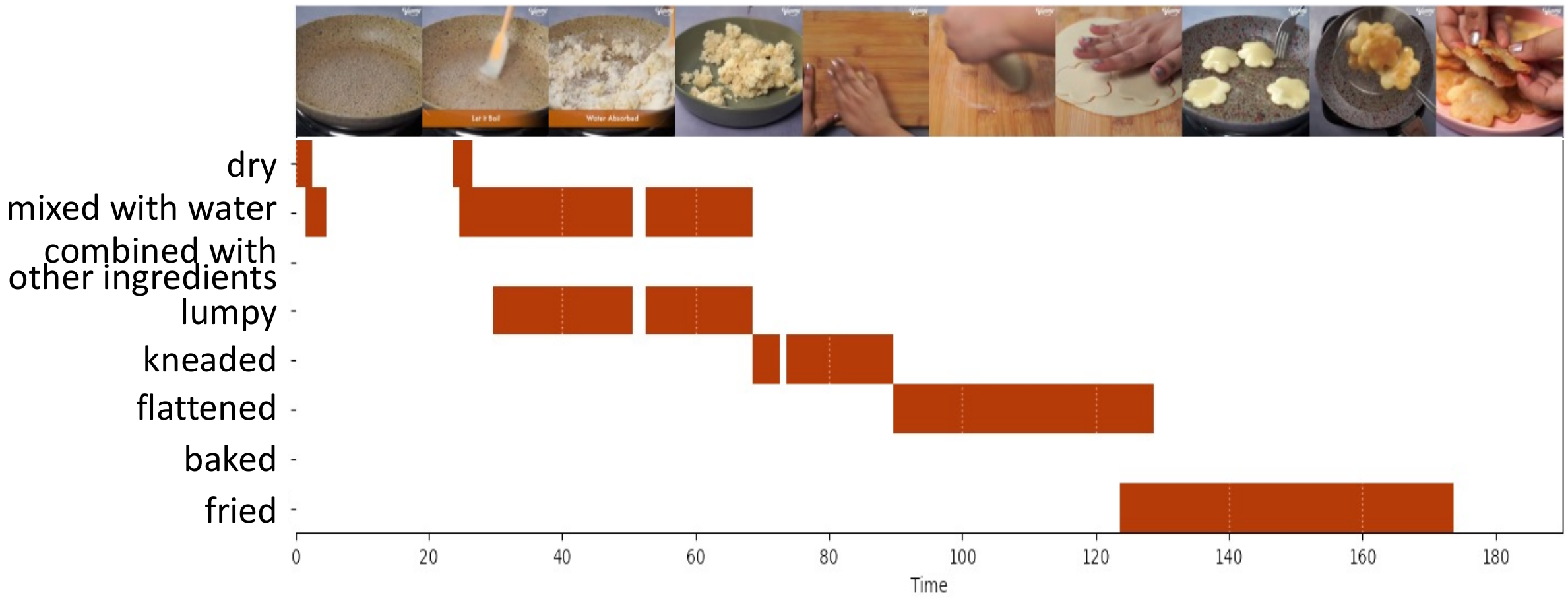}
    \caption{Scene: Cooking food with flour.}
    \vspace{.5cm}
    \label{fig:most_ex_1}
  \end{subfigure}
  \begin{subfigure}{1\linewidth}
    \centering
    \includegraphics[width=1.0\linewidth]{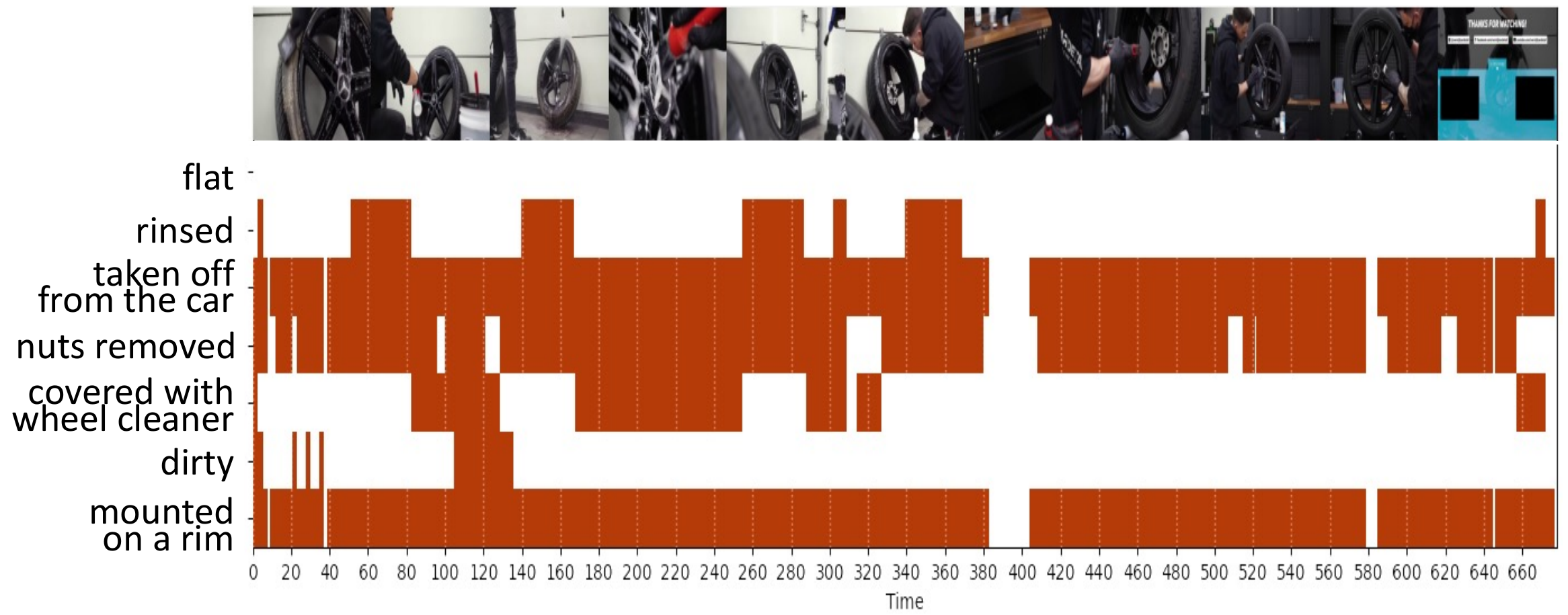}
    \caption{Scene: Washing car tire.}
    \label{fig:most_ex_2}
  \end{subfigure}
  \caption{Annotation example of MOST dataset. Red bands show presence of object states by time (seconds). All labels are marked negative if target objects are not visible.}
  \label{fig:annotation_example}
\end{figure}

As shown in the examples (see Figure~\ref{fig:annotation_example}), we have complicated transition patterns between different states to measure when the object state is actually present.

Although ChangeIt~\cite{look_for_the_change} annotates object state labels as pre/post condition of actions, they only focus on the moment of state change.
For example, the \textit{initial} state of \textit{Cake frosting} is marked right before the action, while distant frames of the unfrosted cake are labeled as \textit{background}, implying object states can’t be identified without state-changing actions.

The MOST dataset covers various object states including those that are not necessarily associated with actions (\eg, straight, dry, flat, smooth), providing a more direct benchmark of object state recognition.

\subsection{Implementation Details}
\label{subsec:implementation}
\noindent \textbf{Training Data}
We used the subset of HowTo100M~\cite{howto100m}, a large-scale collection of YouTube videos, each accompanied by automatically transcribed narrations, as training data.
We assume target object names (\eg, \textit{apple}) and their states to recognize (\eg, \textit{sliced}) are given.
For each object category, we first selected videos that included the object name in the title and narrations.
We further selected videos that included the verbs likely to be associated with the target objects.
We used LLMs to automatically create a list of the likely associated verbs.
For example, from an object-state pair ``peeled apple'', verbs such as ``peel'', ``slice'', and ``cut'' are suggested.
Finally, we removed videos with very long narrations ($>$ 12,000 words) to fit the maximum context length of LLMs, resulting in around 2K videos per object category by adjusting the maximum YouTube search rank (see supplementary for details).
\vspace{0.6em}

\noindent \textbf{Pseudo-Label Generation}
We used the pre-processed narrations from \cite{han2022temporal} that removed low-quality subtitles.
We used GPT-3.5 (\texttt{gpt-3.5-turbo-1106}) for the LLM to process narrations.
Samples that did not follow the specified output format (approximately 1\%) are discarded.
We used LLaVA-1.5 13B \cite{liu2023improved} 4-bit quantized model for VLMs used for temporal alignment.

We applied our framework to videos of 1,123 hours in total.
As a result, pseudo-labels are assigned to 74\% of the video frames.
\vspace{0.6em}

\noindent \textbf{Models}
For the feature backbone, we used the pre-trained InternVideo visual encoder (InternVideo-MM-14-L)~\cite{wang2022internvideo} and sampled 8 frames per second to obtain the feature vectors at 1~fps.
For the {\bf TCN} classifier, we used MS-TCN~\cite{farha2019ms} with 4 stages.
Each stage consists of 10 layers of 512 dimensions, followed by a final linear layer of $K$ outputs with Sigmoid activation.
The {\bf MLP} classifier is a simple MLP consisting of a linear layer of 512 dimensions followed by ReLU activation and the final linear layer of $K$ outputs followed by Sigmoid activation.
The models were trained separately for each object category.
\vspace{0.6em}

\noindent \textbf{Optimization}
We have two learning stages: one for learning the teacher models using pseudo-labels generated by LLMs and the other for self-training with the student model. We train each model with a batch size of 16 for 50 epochs in the first stage, followed by 50 epochs in the second one. We use the Adam optimizer~\cite{kingma2014adam} with decoupled weight decay regularization~\cite{loshchilov2017decoupled}. We set the learning rate to 1e-4 and the weight decay to 0.01.
\vspace{0.6em}

\noindent \textbf{Metrics}
For MOST dataset, we report F1-max and Average Precision (AP) to measure the correctness of temporal localization of object states.
To make scores comparable against uncalibrated zero-shot models, we report F1-max which is the maximum F1 score at the optimal threshold.
We set the thresholds for each object state category.

\subsection{Evaluation on MOST Dataset}
\label{subsec:results_most}

\noindent \textbf{Comparison to off-the-shelf VLMs}
We evaluate our model's multi-label framewise classification peformance, and compare it with three zero-shot VLM models trained from various vision-language data: LLaVA~\cite{liu2023improved}, CLIP~\cite{clip}, and InternVideo~\cite{wang2022internvideo}, to show the effectiveness of learning from explicitly extracted multiple object state information from the narrations. Note that the models from previous works on object state change \cite{joint_discovery, look_for_the_change, multi_task, xue2023learning} are difficult to adapt to our multi-label setting since they assume a single object state change happens only once in a video.

LLaVA is a large multi-modal LLM that accepts a pair of an image and texts.
We used the same model (LLaVA-1.5 13B 4-bit quantized) which is used in interval alignment of our proposed method.
We feed each video frame and the meaning of the object state to LLaVA and prompt it to first describe the image and then answer whether the object state is present in the image.
As the model only produce binary predictions, we only compute the F1-max score.

CLIP~\cite{clip} is a simple yet strong model trained by massive image-text pairs with contrastive learning.
We input each video frame and object state text to the image and text encoder to calculate the similarity score between frames and object states.
Similar to \cite{look_for_the_change}, we tested several text prompts such as ``a photo of \textless object state\textgreater \textless object name\textgreater'' (\eg, ``a photo of cracked egg'') and reported their best results (see supplementary for details).

InternVideo~\cite{wang2022internvideo} is a video-language foundation model based on masked video reconstruction and multi-modal contrastive learning, demonstrating strong results on various video recognition tasks.
We input the model 8 video frames and the object state text similar to CLIP.


Table \ref{tab:comparison_sota} shows the quantitative results.
Our method achieves significantly better results and outperforms all the strong zero-shot baselines in both F1-max and AP, showing the effectiveness of learning from LLM-based object state pseudo-labels.
\vspace{0.6em}

\begin{table*}[tb]
    \centering
    \scalebox{0.9}{
    \begin{tabular}{lcccccccccccccc}
    \toprule
      Method & \multicolumn{2}{c}{Apple} & \multicolumn{2}{c}{Egg} & \multicolumn{2}{c}{Flour} & \multicolumn{2}{c}{Shirt} & \multicolumn{2}{c}{Tire} & \multicolumn{2}{c}{Wire} & \multicolumn{2}{c}{{\bf Average}} \\
      \cmidrule(lr){2-3} \cmidrule(lr){4-5} \cmidrule(lr){6-7} \cmidrule(lr){8-9} \cmidrule(lr){10-11} \cmidrule(lr){12-13} \cmidrule(lr){14-15}
      & F1 & mAP & F1 & mAP & F1 & mAP & F1 & mAP & F1 & mAP & F1 & mAP & F1 & mAP \\
    \midrule      
      LLaVA~\cite{liu2023improved} & 0.34 & \NA & 0.29 & \NA & 0.35 & \NA & 0.28 & \NA & 0.47 & \NA & 0.27 & \NA & 0.33 & \NA \\
      CLIP~\cite{clip} & 0.42 & 0.35 & 0.37 & 0.28 & 0.38 & 0.26 & 0.33 & 0.27 & 0.55 & 0.45 & 0.33 & 0.25 & 0.39 & 0.31 \\
      InternVideo~\cite{wang2022internvideo} & 0.46 & 0.39 & 0.44 & 0.39 & 0.43 & 0.36 & 0.40 & 0.32 & 0.57 & 0.45 & 0.40 & 0.31 & 0.45 & 0.37 \\
      \textbf{Ours} & \textbf{0.53} & \textbf{0.50} & \textbf{0.53} & \textbf{0.48} & \textbf{0.55} & \textbf{0.49} & \textbf{0.50} & \textbf{0.45} & \textbf{0.61} & \textbf{0.52} & \textbf{0.50} & \textbf{0.42} & \textbf{0.54} & \textbf{0.48} \\
  \bottomrule
\end{tabular}
}
    \caption{Comparison with other baselines on MOST dataset. F1 denotes F1-max score and mAP denotes mean Average Precision score averaged across object state categories.}
\label{tab:comparison_sota}
\end{table*}

\begin{figure}[tb]
    \centering
    \includegraphics[width=1\linewidth]{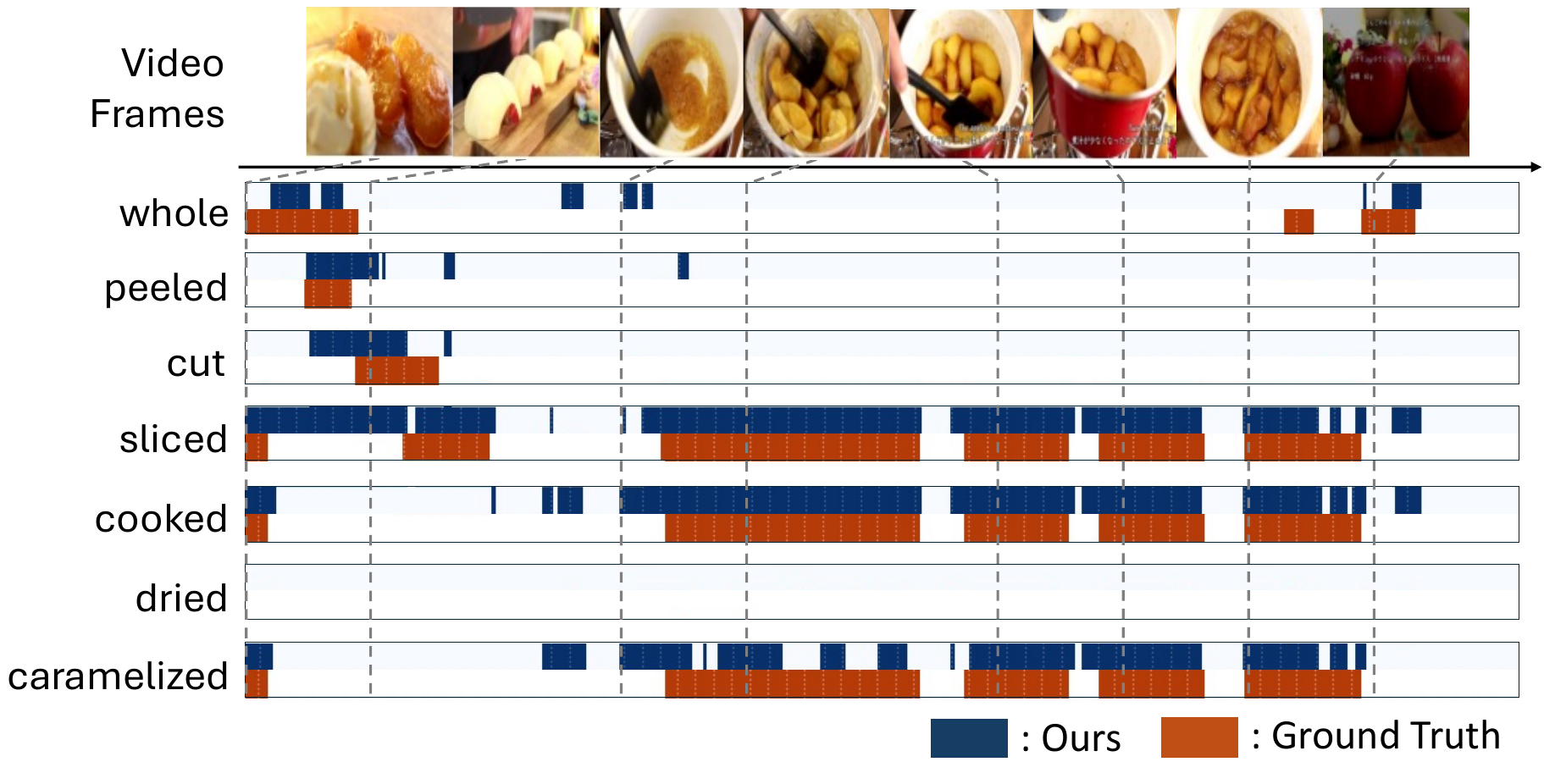}
    \caption{Qualitative results of proposed model (target object: apple). Blue and red bar denote our model's prediction and ground truth, respectively.}
    \label{fig:qualitative}
\end{figure}


\noindent \textbf{Qualitative Analysis}
Figure \ref{fig:qualitative} shows the predictions of our trained model.
Blue and red bar denote the prediction and the ground truth labels, respecitvely. 
Overall our model correctly localize the presence of mulitple object states while there are false positive predictions in some labels (\eg, sliced, cooked).

\subsection{Ablation Study}
\label{subsec:ablation}
\begin{figure}[tb]
  \centering
  \includegraphics[width=1\linewidth]{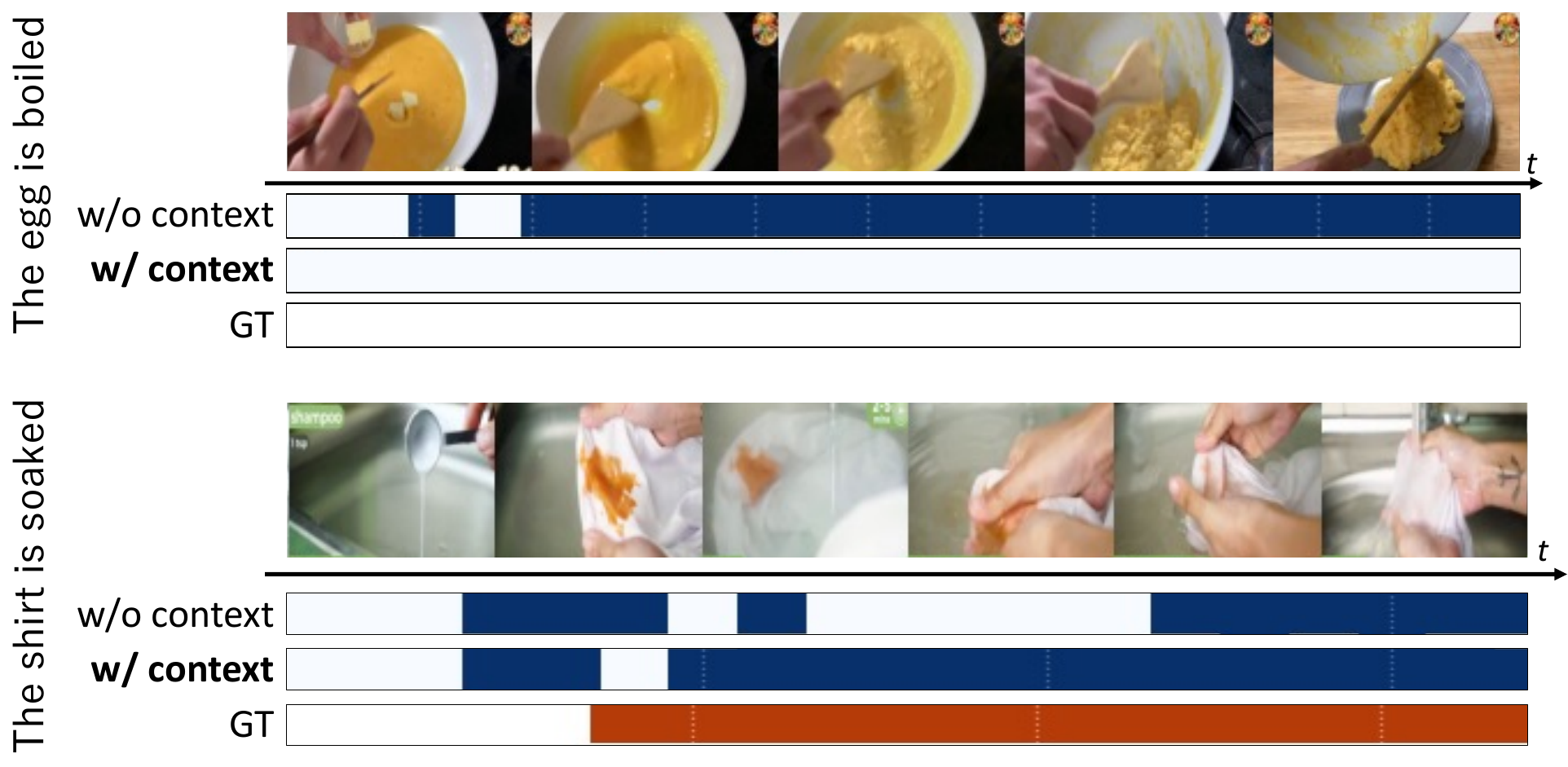}
  \caption{Qualitative analysis on the impact of context-aware pseudo-label generation. Lack of context of state descriptions causes incorrect label assignment (Fig \ref{fig:ablation_no_context}) and subsequently leads to false positive (top) and false negative (bottom) predictions.}
  \label{fig:Qualitative results for ablation study.}
\end{figure}

\noindent \textbf{Does context-aware pseudo-labeling matter?}
We compare the model trained with pseudo-object state labels without using past state context (Table~\ref{tab:ablation} top).
In this model, we infer the state description and the object state labels each from single manipulation action one-by-one.
We observe 3 points gain on mAP, showing that context information matters in inferring object states from actions.
Figure~\ref{fig:Qualitative results for ablation study.} shows that models trained with context-aware labels produce better predictions. 
\vspace{0.6em}

\noindent \textbf{Is self-training effective?}
Table~\ref{tab:ablation} middle shows the performance of {\bf TCN} and {\bf MLP} models directly trained from pseudo-labels.
While {\bf TCN} trained from raw pseudo-labels shows inferior results against {\bf MLP}, its performance surpasses {\bf MLP} by 3 points after self-learning.
\vspace{0.6em}

\noindent \textbf{Does action/state alignment help?}
As shown in Table~\ref{tab:ablation} bottom, action/state alignment using VLMs showed a marginal improvement of 2 points.

\begin{table}[tb]
    \centering
    \begin{tabular}{lcc}
    \toprule
      Method & F1-max & mAP \\
    \midrule
        Full model & \textbf{0.54} & \textbf{0.48}\\
        w/o context-aware pseudo-labels & 0.52 & 0.45\\
        w/o self-training (MS-TCN) & 0.44 & 0.36\\
        w/o self-training (MLP) & 0.51 & 0.45\\
        w/o action/state alignment & 0.53 & 0.46\\
    \bottomrule
    \end{tabular}
    \caption{Ablation results.}
    \label{tab:ablation}
\end{table}

\subsection{Analysis on Generated Pseudo-Labels}

We conduct the quantitative and qualitative analysis of generated pseudo-labels to validate that our context-aware framework has a positive effect on generating accurate labels. We sample 55 videos from the training set (approx. nine videos per category) and manually annotate them to create a sub-dataset for pseudo-label evaluation. Note that we don't tune the training using these annotations.

We compare our pseudo-labeling framework with zero-shot baselines as described in \ref{subsec:results_most}. We also ablate the context awareness in our method to demonstrate its impact. We use F1@\{10, 20, 30\} score averaging over all object categories, corresponding to thresholds of 0.1, 0.2, and 0.3, respectively, based on CLIP and InternVideo score distributions. These thresholds are used due to the lack of ground truth for optimal pseudo-labeling thresholds. Both our method and LLaVA~\cite{liu2023improved} consistently output binary predictions, maintaining stable performance across thresholds.

The Table \ref{tab:pseudo-label_f1} shows the results. Our method outperforms all baselines, demonstrating the effectiveness of using LLMs to extract object state information from the video narrations. Comparing our method with and without the context shows the impact of past context, consistent with the ablation study in Section \ref{subsec:ablation}.

\begin{figure*}[tb]
    \centering
    \includegraphics[width=0.9\linewidth]{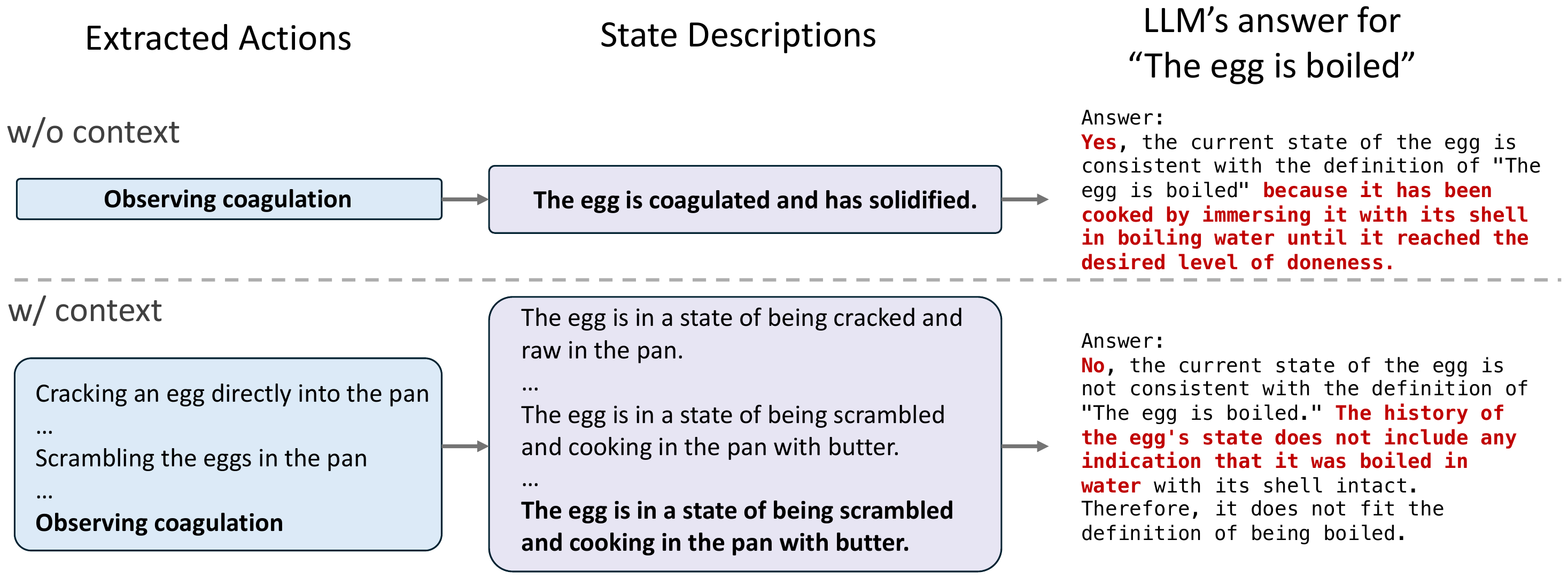}
    \caption{Effect of using past context on object state label inference. Action ``observing coagulation'' tells only that egg is changing from unsolidified to solidified state, but it lacks other information such as whether egg is in shell or broken. These lack of information often leads to incorrect reasoning (top). This issue is mitigated by considering the influence of past states by incorporating past actions and state descriptions (bottom).}
    \label{fig:ablation_no_context}
\end{figure*}

\begin{table}[tb]
    \centering
    \begin{tabular}{lcccc}
    \toprule
      Method & F1@10 & F1@20 & F1@30 \\
    \midrule
        Ours (LLM) & \textbf{0.35} & \textbf{0.35} & \textbf{0.35} \\
        Ours (LLM) w/o context & 0.31 & 0.31 & 0.31 \\
        LLaVA~\cite{liu2023improved} & 0.26 & 0.26 & 0.26 \\
        CLIP~\cite{clip} & 0.22 & 0.23 & 0.00 \\      InternVideo~\cite{wang2022internvideo} & 0.20 & 0.23 & 0.02 \\
        
    \bottomrule
    \end{tabular}
    \caption{Comparison of F1@\{10, 20, 30\} scores of pseudo-labels generated by our framework and other baselines. Since our method and LLaVA~\cite{liu2023improved} produce binary outputs for pseudo-labels, scores of F1@\{10, 20, 30\} are all same.}
    \label{tab:pseudo-label_f1}
\end{table}


Figure \ref{fig:ablation_no_context} illustrates how our context-aware framework improves object state label accuracy. By leveraging the whole past context, state descriptions can better represent the egg's state at each moment and capture the influence between different states (how \textit{broken} state of an egg has an influence on becoming \textit{solidified} state).

Typical failure cases occur when our framework fails to extract the correct actions from the video due to unclear narrations, leading to inaccurate object state inference. While incorporating VLMs to fill in missing actions could be a potential solution, VLMs might also hallucinate non-existent actions, negatively impacting the pseudo-labels. We leave the issue of incorrect action detection for future work, as our current focus is on learning object states.



\subsection{Evaluation on ChangeIt Dataset}
\label{subsec:results_changeit}
ChangeIt dataset \cite{look_for_the_change} provides annotations for \textit{initial state}, \textit{action}, \textit{end state}, and \textit{background} labels for each video, aiming to evaluate the temporal localization of the first three labels. For 44 state-changing actions, 667 videos up to 48 hours are annotated for evaluation. 

For training, we use a subset of ChangeIt training set with narrations for pseudo-label generation, selecting at most 350 videos per category with the highest noise adaptive weight~\cite{look_for_the_change}, resulting in 10,749 videos (32\% of the original training set). We train all baselines using this subset. We also modified our pseudo-labeling framework to predict \textit{initial state}, \textit{action}, or \textit{end state}. we assign the \textit{background} label by leveraging CLIP~\cite{clip} to calculate the similarity score between the object name and the video frames. Due to high computational demands, we omit the matching between state description and video frames.


We optimize the model as described in Section \ref{subsec:implementation} and apply causal ordering constraints~\cite{look_for_the_change} during inference, selecting frames with the highest probabilities while preserving causal order for fair comparison.



Following \cite{look_for_the_change} we use precision@1 for metrics, but only report state precision as we focus on state recognition.


Table~\ref{tab:comparison_changeit} shows the results.
Our method achieves better results than other baselines without relying on heuristic constraints during training, demonstrating its flexibility to work effectively in different settings.

\begin{table}[tb]
  \centering
  \begin{tabular}{p{4.0cm}c}

    \toprule
    Method &  State prec.\\
    \midrule
    Random w/ constraint & 0.15 \\
    Look for the Change \cite{look_for_the_change} & 0.23 \\ 
    Multi-Task  \cite{multi_task} &  0.38\\ 
    Xue et al~\cite{xue2023learning} & 0.41\\ 
    \textbf{Ours} & \textbf{0.42} \\ 
   \bottomrule
  \end{tabular}
   \caption{Comparison on ChangeIt dataset, showing precition@1 for state recognition.}  
\label{tab:comparison_changeit}
\end{table}

\section{Conclusion}
\label{sec:conclusion}

We have developed a new framework for learning multiple object states from narrated videos leveraging LLMs.
The key idea is to infer the existence of object states from actions included in video narrations while considering the influence of past states with the help of LLMs.
We have shown that learning from explicitly generated pseudo-labels demonstrates significant improvement against strong vision-language models.
We have further analyzed that considering the past context of object states boosts the accuracy of pseudo-labels, improving the trained model.
This work opens up the possibility of using LLMs as a catalyst for recognizing label-scarce concepts through label-abundant concepts.

\noindent
\textbf{Acknowledgement} This work was supported by JST ASPIRE Grant Number JPMJAP2303 and JSPS KAKENHI Grant Numbers
JP23H00488 and JP24K02956.


{\small
\bibliographystyle{ieee_fullname}
\bibliography{egbib}
}

\clearpage
\setcounter{page}{1}
\setcounter{figure}{0}
\setcounter{table}{0}

\appendix
\addcontentsline{toc}{section}{Appendix} 
\part{Appendix} 
\parttoc 

\section{Details on MOST Dataset}
\label{supsec:most_dataset}
\subsection{Object State Category and Their Descriptions}
MOST dataset covers six object categories: apple, egg, flour, shirt, tire, and wire. This selection covers diverse situations such as cooking, housework, hobbies, cars, crafts, and electronics. We select around 10 object states per object category and provide a description for each category name to ensure consistent annotation across the dataset. The description essentially follows the dictionary meaning, but to avoid ambiguity, especially regarding the conditions under which a state disappears, these conditions are explicitly included in the description.
Table~\ref{tab:category_definition_apple} - \ref{tab:category_definition_wire} shows the full list of object categories, state names, and their descriptions.
This information is used for annotation, pseudo-label generation, and model training/inference.

\begin{table*}[t]
    \centering
    \begin{tabular}{|l|p{2cm}|p{12cm}|}
    \hline
    Object & State & description \\
    \hline
\multirow{11}{*}{apple}
 & whole & \small{ "The apple is whole" refers to the state of an apple that has not been cut or sliced and retains its original round shape. Even if it has been peeled or cooked, it is still applicable as long as it retains its original round shape.} \\
\cline{2-3}
 & peeled & \small{"The apple is peeled" refers to the state of an apple that has had its skin removed by peeling. If the apple is cooked or loses its round shape by being sliced or grated, we don't say it is peeled.} \\
\cline{2-3}
 & grated & \small{"The apple is grated" refers to the state of an apple that has been shredded or grinded into finer or smoother pieces. Sometimes grating can release moisture, but 'grated' refers to the presence of coarse grains. It no longer applies to smooth smoothies or sauces made by mixers or blenders.} \\
\cline{2-3}
 & cut & \small{"The apple is cut" refers to the state of an apple that has been coarsely split into 2 or 3 pieces. If it is sliced into small pieces, we don't say it is cut. It doesn't matter whether the apple is peeled or not.} \\
\cline{2-3}
 & sliced & \small{"The apple is sliced" refers to the state of an apple that has been cut into thin or narrow pieces. It does not include the case where the apple is split into two or three large pieces. Even if it is cooked, it is still applicable as long as it retains the shape of the pieces.} \\
\cline{2-3}
 & cooked & \small{"The apple is cooked" refers to the state of an apple that has been sufficiently heated in any way.} \\
\cline{2-3}
 & baked & \small{"The apple is baked" refers to the state of an apple that has been cooked in an oven.} \\
\cline{2-3}
 & smooth & \small{"The apple is smooth" refers to the state of an apple that has been squeezed or blended into a smooth texture. If the apple has turned to a liquid(completely juiced), we don't say it is smooth.} \\
\cline{2-3}
 & dried & \small{"The apple is dried" refers to the state of an apple that has been dried by removing moisture. This can be done by various methods, such as drying in the sun, drying in the oven, or drying in a dehydrator.} \\
\cline{2-3}
 & mixed with other ingredients & \small{"The apple is mixed with other ingredients" refers to the state of apple combined with other components in a recipe. It is applicable only when we can see multiple ingredients are mixed. If they have dissolved into a single object, the state doesn't apply. It doesn't refer to adding only seasonings or water. This state is not affected by the fact it is cooked or not.} \\
\cline{2-3}
 & caramelized & \small{"The apple is caramelized" refers to the state of apple that has been cooked with brown sugar, typically until it turns a golden brown color.} \\
 \hline
    \end{tabular}
    \caption{Object State Category and description for Apple}
    \label{tab:category_definition_apple}
\end{table*}
\begin{table*}[t]
    \centering
\begin{tabular}{|l|p{2cm}|p{12cm}|}
    \hline
    Object & State & description \\
    \hline
\multirow{13}{*}{egg}
 & intact & \small{"The egg is intact" refers to the state of an egg's shell being unbroken or undamaged in a cooking situation. It implies that the egg has not cracked or shattered, ensuring that the contents of the egg, remain contained within the shell. As long as the shell is unbroken, the state applies even if it is boiled.} \\
\cline{2-3}
 & raw & \small{"The egg is raw" refers to an egg that has not been heated. It means that the egg's contents, including the yolk or egg white, have not undergone any cooking process involving heating, such as boiling, frying, or baking. As long as the egg has not been heated, it is raw even if it is whisked or mixed with other ingredients.} \\
\cline{2-3}
 & boiled & \small{"The egg is boiled" refers to the state of an egg cooked by immersing it with its shell in boiling water until it reaches the desired level of doneness. If the egg is boiled with its shell cracked open, "The egg is boiled" doesn't apply because it is poached.} \\
\cline{2-3}
 & poached & \small{"The egg is poached" refers to the state in which the egg is cooked by cracking it into hot water. It does not apply if the egg is boiled with the shell still attached.} \\
\cline{2-3}
 & cracked & \small{"The egg is cracked" refers to the state of a broken eggshell. After the eggshell is disposed of or removed, it is not "cracked" anymore.} \\
\cline{2-3}
 & runny & \small{"The egg is runny" refers to the state of the egg yolk or the entire egg when it is not fully cooked and has a semi-liquid consistency. This applies only if the yolk can be seen to be thick and liquid. If the egg is raw or cracked where it is not heated at all, "The egg is runny" doesn't apply.} \\
\cline{2-3}
 & whisked & \small{"The egg is whisked" refers to state of an egg beaten using a whisk or fork until the yolk and white are fully combined. The statement includes when only yolk or white is whisked. If the egg has been heated and cooked, "The egg is whisked" doesn't apply.} \\
\cline{2-3}
 & cooked & \small{"The egg is cooked" refers to the state of an egg when it has been heated sufficiently to reach a desired level of doneness. This doesn't apply to raw eggs or eggs that have been heated but are still runny.} \\
\cline{2-3}
 & fried & \small{"The egg is fried" refers to the state of an egg cooked by cracking it into a heated pan with oil or butter until the white part is cooked and the yolk is still runny or fully cooked. To be "fried" a certain degree of dry texture is necessary, and watery scrambled eggs is not applicable.} \\
\cline{2-3}
 & scrambled & \small{"The egg is scrambled" refers to the state of the cooked egg where the yolk and white of the egg are mixed together and cooked until they form a soft, creamy texture. "The egg is scrambled" applies if the egg has formed a texture by heat. If the egg is not heated well, the state is not applicable.} \\
\cline{2-3}
 & foamy & \small{"The egg is foamy" refers to the state of beaten egg whites that have been whisked vigorously until they form a light and airy texture. This is achieved by incorporating air into the egg whites, resulting in a foam-like consistency with soft peaks. If the egg has been cooked, the statement doesn't apply.} \\
\cline{2-3}
 & baked & \small{"The egg is baked" refers to the state of an egg cooked by placing it in an oven or similar heat source. Baking an egg typically involves cracking it into a dish or ramekin and then cooking it until the egg white is set and the yolk is still slightly runny.} \\
\cline{2-3}
 & mixed with other ingredients & \small{"The egg is mixed with other ingredients" refers to the state of eggs combined with other components in a recipe. It is applicable only when we can see multiple ingredients are mixed. The mix of only egg white and egg yolk is not applicable here. If they have dissolved into a single object, the state doesn't apply. It doesn't refer to adding only seasonings or water. This state is not affected by the fact it is cooked or not.} \\
\hline
\end{tabular}
\caption{Object State Category and description for Egg}
\label{tab:category_definition_egg}
\end{table*}
\begin{table*}[t]
    \centering
\begin{tabular}{|l|p{2cm}|p{12cm}|}
    \hline
    Object & State & description \\
    \hline
\multirow{8}{*}{flour}
 & dry & \small{"The flour is dry" refers to the state of flour when it has not absorbed any moisture or liquid. In this condition, the flour remains in its loose, powdery form, without any clumps or dampness. If any moisture like water or egg is added, it is no longer dry.} \\
\cline{2-3}
 & mixed with water & \small{"The flour is mixed with water" refers to the state of flour that has been combined with water. If it is sufficiently kneaded or cooked, losing the fluidity, "The flour is mixed with water" is no longer applicable.} \\
\cline{2-3}
 & combined with other ingredients & \small{"The flour is combined with other ingredients" refers to the state of flour that has been combined with other ingredients in a recipe. It doesn't refer to adding only seasonings or water.} \\
\cline{2-3}
 & lumpy & \small{"The flour is lumpy" refers to the state of flour when it contains small clumps or aggregates, often as a result of moisture exposure or poor storage conditions. If the flour is sufficiently kneaded to remove the clumps, "The flour is lumpy" is no longer applicable.} \\
\cline{2-3}
 & kneaded (dough) & \small{"The flour is kneaded dough" refers to the state of flour that has been kneaded by hand or with a machine. Once the dough is flattened or cooked, "The flour is kneaded dough" is no longer applicable.} \\
\cline{2-3}
 & flattened & \small{"The flour is flattened" refers to the state of flour that it has been kneaded and stretched thin, or has become pacake butter and is spread thin. If it is not flattened, "The flour is flattened" is not applicable.} \\
\cline{2-3}
 & baked & \small{"The flour is baked" refers to the state of flour that has undergone a cooking process, typically in an oven or another heat source, which transforms its raw form into a firmer, cohesive and often golden-brown structure. If it is cooked in panful of oil, it is not baked but fried.} \\
\cline{2-3}
 & fried & \small{"The flour is fried" refers to the state of flour that has been cooked in panful of oil until it changes into crispy texture and often golden-brown color.} \\
\hline
\end{tabular}
\caption{Object State Category and description for Flour}
\label{tab:category_definition_flour}
\end{table*}
\begin{table*}[t]
    \centering
\begin{tabular}{|l|p{2cm}|p{12cm}|}
    \hline
    Object & State & description \\
    \hline
\multirow{10}{*}{shirt}
 & stained & \small{"The shirt is stained" refers to the state of the shirt that has been marked or discolored by a foreign substance. This could have been caused by a variety of factors such as spilled food, ink, dirt, sweat, or any other substance that might leave a residue or color behind. However, if the color is intentionally added (e.g. dyeing), "The shirt is stained" does not apply.} \\
\cline{2-3}
 & soaked & \small{"The shirt is soaked" refers to the state of the shirt that has absorbed a significant amount of liquid, rendering it wet throughout. In this context, "soaked" does not merely mean a small splash or a few droplets of water but indicates that the shirt is thoroughly drenched.} \\
\cline{2-3}
 & dry & \small{"The shirt is dry" refers to the state of the shirt that indicates an absence of moisture or wetness on or within the fabric.} \\
\cline{2-3}
 & wrinkled & \small{"The shirt is wrinkled" refers to the state of the shirt that exhibits creases, folds, or irregularities on its surface rather than being smooth or flat.} \\
\cline{2-3}
 & smooth & \small{"The shirt is smooth" refers to the state of the shirt that indicates it has a flat and even surface without any wrinkles, creases, or rough patches.} \\
\cline{2-3}
 & dyed & \small{"The shirt is dyed" refers to the state of the shirt that has undergone a process to change its color. In this context, 'dyed' means the shirt has been treated with dyes or pigments to impart a specific color or design onto its fabric.} \\
\cline{2-3}
 & folded & \small{"The shirt is folded" refers to the state of the shirt that has been folded to make it more compact, neat, and easier to store.} \\
\cline{2-3}
 & piled & \small{"The shirt is piled" refers to the state of a shirt that has been stacked or heaped with other items, possibly other shirts or clothes.} \\
\cline{2-3}
 & rolled & \small{"The shirt is rolled" refers to the state of the shirt that has been folded or twisted into a cylindrical or tubular shape by rolling.} \\
\cline{2-3}
 & hung & \small{"The shirt is hung" refers to the state of the shirt that indicates it has been suspended, usually from a hanger, hook, or another support structure.} \\
\hline
\end{tabular}
\caption{Object State Category and description for Shirt}
\label{tab:category_definition_shirt}
\end{table*}
\begin{table*}[t]
    \centering
\begin{tabular}{|l|p{2cm}|p{12cm}|}
    \hline
    Object & State & description \\
    \hline
\multirow{7}{*}{tire}
 & flat & \small{"The tire is flat" refers to the state of a tire that has lost its internal air pressure (deflated) to the point where it no longer maintains its rounded shape or provides the necessary cushioning for a vehicle's load.} \\
\cline{2-3}
 & rinsed & \small{"The tire is rinsed" refers to the state of a tire that has been washed away by water. If the soap or cleaner is on the tire, "The tire is rinsed" does not apply.} \\
\cline{2-3}
 & taken off from the car & \small{"The tire is taken off from the car" refers to the state of the tire that is no longer attached or mounted to the vehicle.} \\
\cline{2-3}
 & nuts removed & \small{"The tire is nuts removed" refers to the state of a tire that has had its lug nuts or wheel nuts removed. These nuts are used to secure the wheel to the vehicle's axle, ensuring that the tire stays in place while the vehicle is in motion.} \\
\cline{2-3}
 & covered with wheel cleaner & \small{"The tire is covered with wheel cleaner" refers to the state of the tire that has been treated or coated with a specific cleaning foam designed for wheels. If we cannot see the foam, "The tire is covered with wheel cleaner" does not apply.} \\
\cline{2-3}
 & dirty & \small{"The tire is dirty" refers to the state of the tire that indicates contamination or uncleanliness with soil on its surface.} \\
\cline{2-3}
 & mounted on a rim & \small{"The tire is mounted on a rim" refers to the state of a tire that has been attached and secured onto a wheel's metal structure, commonly known as the rim. This condition applies from the point where the tire is fully engaged in the rim.} \\
\hline
\end{tabular}
\caption{Object State Category and description for Tire}
\label{tab:category_definition_tire}
\end{table*}
\begin{table*}[t]
    \centering
\begin{tabular}{|l|p{2cm}|p{12cm}|}
    \hline
    Object & State & description \\
    \hline
\multirow{10}{*}{wire}
 & straight & \small{"The wire is straight" refers to the state of a straight wire or cable that has not been coiled or twisted. A wire can become straight by pulling it and add tension to it.} \\
\cline{2-3}
 & twisted & \small{"The wire is twisted" refers to the state of a wire or cable that has been rotated around itself or another wire, causing it to form a helical or spiral shape.} \\
\cline{2-3}
 & soldered & \small{"The wire is soldered" refers to the state of a wire that has been joined or attached to another component using solder. If the solder is not used, it is not soldered.} \\
\cline{2-3}
 & cut & \small{"The wire is cut" refers to the state of a wire or cable that has been cut using a tool such as a wire cutter. If the end of cut wire is connected to the other end or to another component, it is not cut anymore.} \\
\cline{2-3}
 & damaged & \small{"The wire is damaged" refers to the state of a wire or cable that has been damaged or broken. If the wire is intentionally cut, stripped, or soldered, it is not considered damaged.} \\
\cline{2-3}
 & exposed & \small{"The wire is exposed" refers to the state of a wire or cable with its skin or insulation removed, exposing the metal underneath. If the exposed wire is connected to another component, it is not exposed anymore.} \\
\cline{2-3}
 & coiled & \small{"The wire is coiled" refers to the state of a coiled wire or cable that has been wrapped around itself or another wire, causing it to form a helical or spiral shape. We don't say it is coiled if it has only one loop.} \\
\cline{2-3}
 & braided & \small{"The wire is braided" refers to the state of a wire that has been woven or interlaced into a braided pattern.} \\
\cline{2-3}
 & looped & \small{"The wire is looped" refers to the state of a wire that has been bent into a loop or circle. If wires are tangled, it is not looped. If a wire is not looped if it is coiled.} \\
\cline{2-3}
 & connected & \small{"The wire is connected" refers to the state of a wire or cable that has been attached to another component or element (not wire). If the wire is connected with another wire, we don't say it is not connected.} \\
\hline
\end{tabular}
\caption{Object State Category and description for Wire}
\label{tab:category_definition_wire}
\end{table*}
 
\subsection{Ground Truth Label Annotation}
\begin{figure*}[t]
  \centering
  
  \begin{subfigure}{.32\textwidth}
    \centering
    \includegraphics[width=\linewidth]{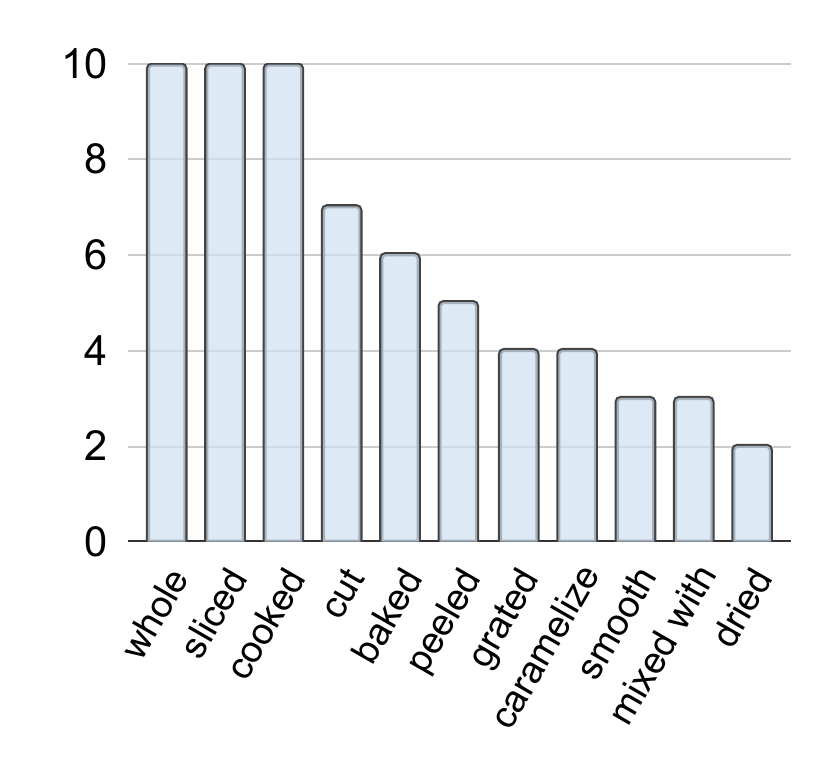}
    \caption{Apple}
    \label{fig:annot_apple}
  \end{subfigure}
  \hfill
  \begin{subfigure}{.32\textwidth}
    \centering
    \includegraphics[width=\linewidth]{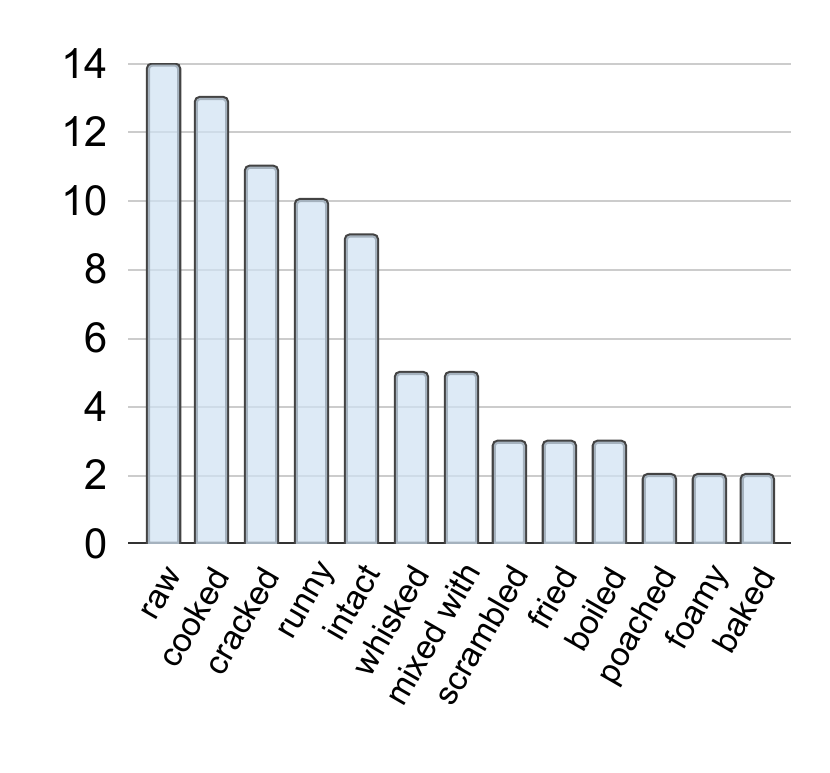}
    \caption{Egg}
    \label{fig:annot_egg}
  \end{subfigure}
  \hfill
  \begin{subfigure}{.32\textwidth}
    \centering
    \includegraphics[width=\linewidth]{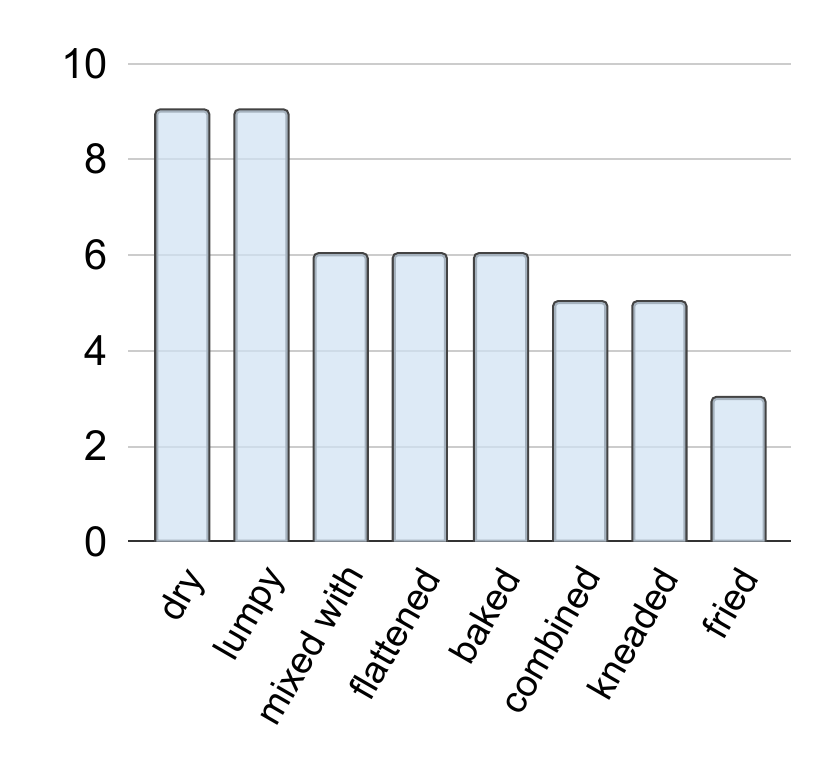}
    \caption{Flour}
    \label{fig:annot_flour}
  \end{subfigure}
  
  \begin{subfigure}{.32\textwidth}
    \centering
    \includegraphics[width=\linewidth]{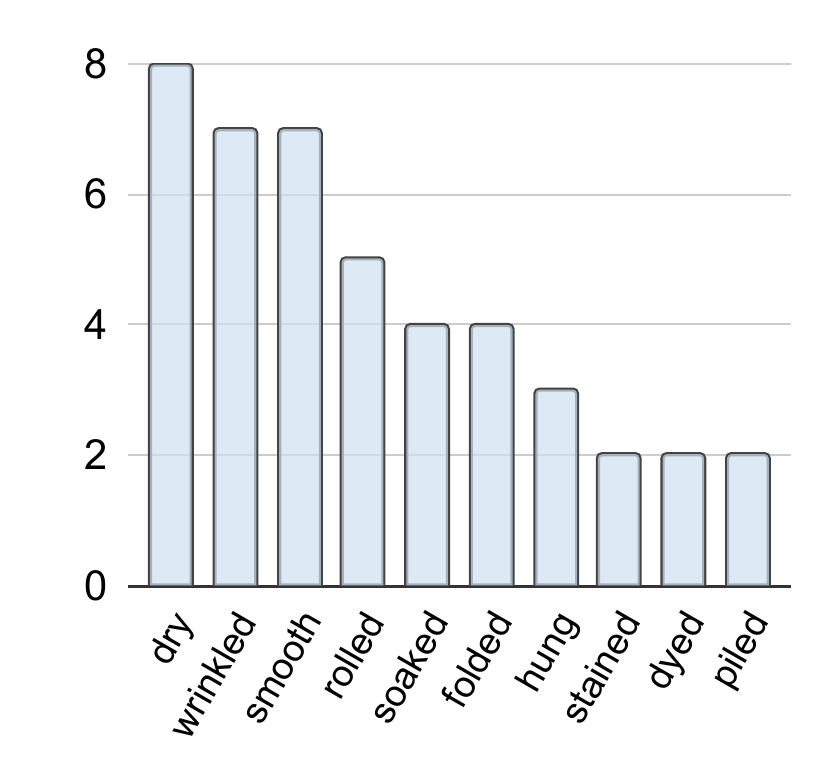}
    \caption{Shirt}
    \label{fig:annot_shirt}
  \end{subfigure}
  \hfill
  \begin{subfigure}{.32\textwidth}
    \centering
    \includegraphics[width=\linewidth]{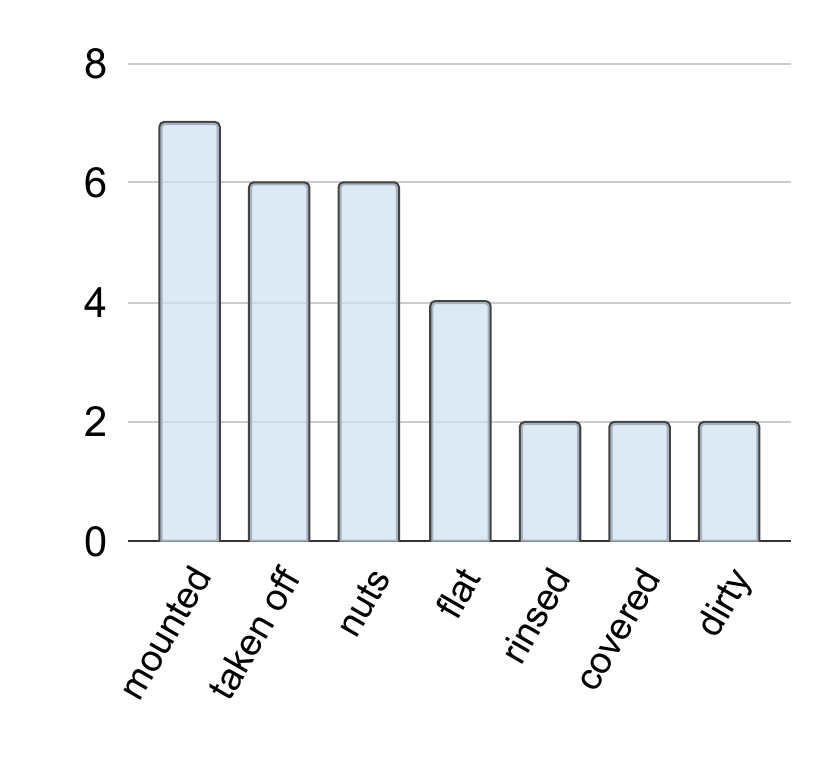}
    \caption{Tire}
    \label{fig:annot_tire}
  \end{subfigure}
  \hfill
  \begin{subfigure}{.32\textwidth}
    \centering
    \includegraphics[width=\linewidth]{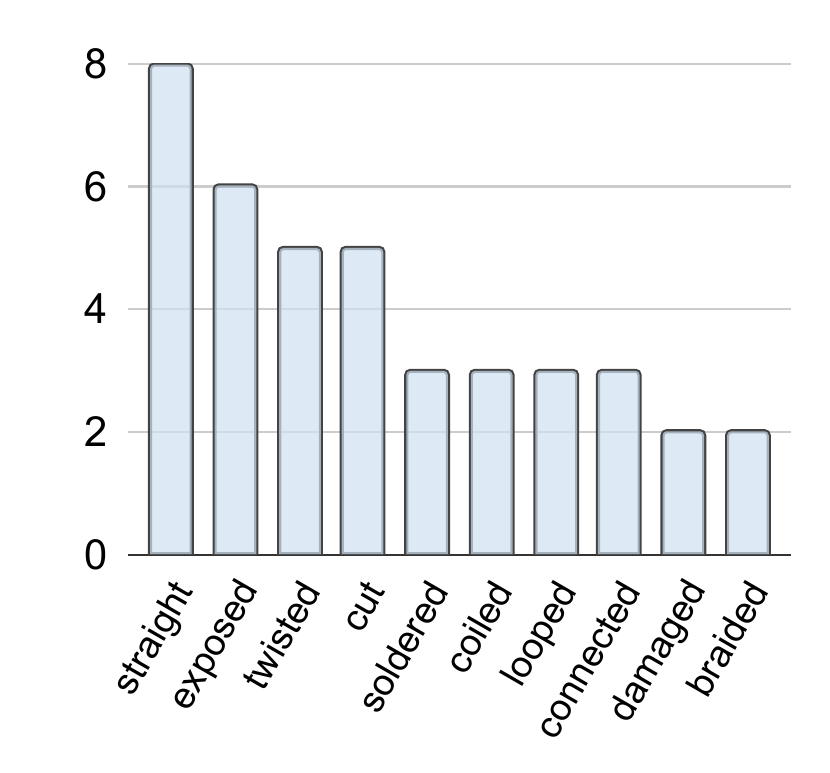}
    \caption{Wire}
    \label{fig:annot_wire}
  \end{subfigure}
  
  \caption{Data distribution of MOST dataset. Y-axis denotes number of videos that each state presents.}
  \label{fig:most_statistics}
\end{figure*}

For evaluation, we manually collected YouTube videos for each object category and densely annotated the interval of object states' presence at 30~fps based on the object state names and their description.
As a result, 61 videos with a total duration of 159.6 minutes were fully annotated.
On average, five different states appeared in a single video. This is more than that of ChangeIt \cite{look_for_the_change} dataset where only \textit{initial}/\textit{end} states are present in one video. Figure \ref{fig:most_statistics} provides the detailed distribution of MOST dataset.

During annotation, trained annotators were asked to annotate object states based on the description.
If multiple object instances in different states appear in one frame, all of the present state labels are assigned to the frame. For example, if two apples are present in one frame, one peeled but the other sliced, the ``peeled'' and the ``sliced'' labels are assigned to the frame.

\clearpage
\begin{table}[t]
    \centering
    \begin{tcolorbox}
    
    \small{
    Which action most describes the image? Choose from the options below. The answer should be taken verbatim from the text of the option. \\

    \textcolor{blue}{\{list of candidate actions\}}
    }
    \end{tcolorbox}
    \caption{Prompt to choose most plausible action. We replace \textcolor{blue}{\{list of candidate actions\}} with candidate actions listed with ``-'' as first character and concatenated by newline. Black text is used as is.}
    \label{tab:prompt_llava_action}
\end{table}
\begin{table}[t]
    \centering
    \begin{tcolorbox}
    
    \small{
    The image possibly shows people \textcolor{blue}{\{inferred action\}}.
    Describe the progress of the action in detail. And then answer whether ``\textcolor{blue}{\{state\}}" is true or not. Finally, specify the judgement with "The answer is True/False".
    }
    \end{tcolorbox}
    \caption{Prompt to filter unrelated video frame. We replace \textcolor{blue}{\{inferred action\}} with selected plausible action, and \textcolor{blue}{\{state\}} with object state text (\eg \textit{The egg is raw}) . Black text is used as is.}
    \label{tab:prompt_llava_state}
\end{table}

\section{Full Prompts Used to Infer Object States via LLM}
\label{supsec:full_prompt_for_llm}
We proposed a method to infer object state labels from the narrations of videos in Section 3.2 of the main paper. The process consists of (a) extracting manipulation actions from narrations, (b) generating state descriptions from the actions, and (c) inferring context-aware object state labels. Table \ref{tab:prompt_a}, \ref{tab:prompt_b}, and \ref{tab:prompt_c} show the prompts for (a), (b), and (c), respectively.

For each process, outputs from the LLM are parsed and we use only direct answers with a specified format. See Section \ref{subsubsec:llm_ex} for examples of raw LLM output.
\begin{table}[t]
    \centering
    \begin{tcolorbox}
    \small{
    Analyze a segment of video transcript provided in CSV format. The CSV only has one column and no headers. \\
\textcolor{blue}{\{block of transcribed narrations\}} \\

You need to list and describe all object manipulating actions performed in the video in detail. Do not include actions such as greeting, thanking, explaining or summarizing that do not manipulate any object. Do not summarize actions too short, but make sure you describe all the actions in each sentence in detail. Especially, make sure to use original nouns (object names) and verbs (human actions) when you summarize.\\

In addition, for each action, extract the part of the transcript that describes or supports the action. Make sure to extract the whole sentence for support.
When you need to combine multiple lines from the transcript to support an action, separate them with a space instead of a comma or line break.\\

The answer format should be in CSV format. Make sure to use quotation marks for each action and the part of the transcript.\\
Format: "$<$detailed summary of action$>$","$<$part of the transcript (This should be exactly the same as the original. Don't skip.)$>$"\\
Example: "Adding whisked eggs into the pan.","let's add the whisked eggs into the pan"}
    \end{tcolorbox}
    \caption{Prompt for manipulation action extraction. We replace \textcolor{blue}{\{block of transcribed narrations\}} with at most 10 sentences of transcribed narrations concatenated by newline. Black text is used as is.}
    \label{tab:prompt_a}
\end{table}
\begin{table}[t]
    \centering
    \begin{tcolorbox}
    \small{
    You will be given a sequence of actions.
Trace the history of changes in the internal state of \textcolor{blue}{\{object\}} and describe it in detail for each action. \\
The initial state of the \textcolor{blue}{\{object\}} is "\textcolor{blue}{\{previous state description\}}". You don't need to include the initial state in the answer.\\

The answer format should be in CSV with the action column and state description column.
Make sure that each state description includes the whole history of what has been done on the \textcolor{blue}{\{object\}} so far. The description should be a complete sentence starting with "The \textcolor{blue}{\{object\}}", but do not finish only with this.\\
If the internal state doesn't change after the action, you don't have to change the state description from the previous one. Use quotation marks for the description.
The answer format:\\
"action","state"\\

Here is the sequence of actions.\\
\textcolor{blue}{\{action sequence\}}
    }
    \end{tcolorbox}
    \caption{Prompt to generate object state description. We replace \textcolor{blue}{\{object\}} with primary object name (Table \ref{tab:object_names}), \textcolor{blue}{\{previous state description\}} with previous last state description, \textcolor{blue}{\{action sequence\}} with 10 actions concatenated by newline. Black text is used as is.}
    \label{tab:prompt_b}
\end{table}
\begin{table}[t]
    \centering
    \begin{tcolorbox}
    \small{
    This is a history of state of \textcolor{blue}{\{object\}}:\\
\textcolor{blue}{\{list of state description up to that point\}}\\

Now, does the state of \textcolor{blue}{\{object\}} fit the definition of "\textcolor{blue}{\{state\}}"?\\

Object state definition:\\
\textcolor{blue}{\{description\}}\\

Think step-by-step as follows:\\
- First, list all points for judging the state "\textcolor{blue}{\{state\}}" from the object state definition. Make sure to describe in detail.\\
- Second, carefully compare all listed judging points to the whole history of the object state by tracing it in detail.\\
- Then, answer Yes/No about whether the current state of \textcolor{blue}{\{object\}} is consistent with the definition and give detailed reasons. If the history doesn't contain enough information for judging, answer Ambiguous.\\

Make sure to answer the three things above in detail, separating them by newline as follows:\\
Judging points: [judging points from object state definition]\\

Comparison: [comparison]\\

Answer: [yes/no/ambiguous and why]
    }
    \end{tcolorbox}
    \caption{Prompt to infer object state label from descriptions. We replace \textcolor{blue}{\{object\}} with primary object name (Table \ref{tab:object_names}), \textcolor{blue}{\{list of state description up to that point\}} with state descriptions concatenated by newline, \textcolor{blue}{\{state\}} with object state text (\eg \textit{The egg is raw}), and \textcolor{blue}{\{description\}} with description of state. Black text is used as is.}
    \label{tab:prompt_c}
\end{table}

\section{Additional Information on Interval Alignment}
\begin{figure*}[tb]
    \centering
    \includegraphics[width=1\linewidth]{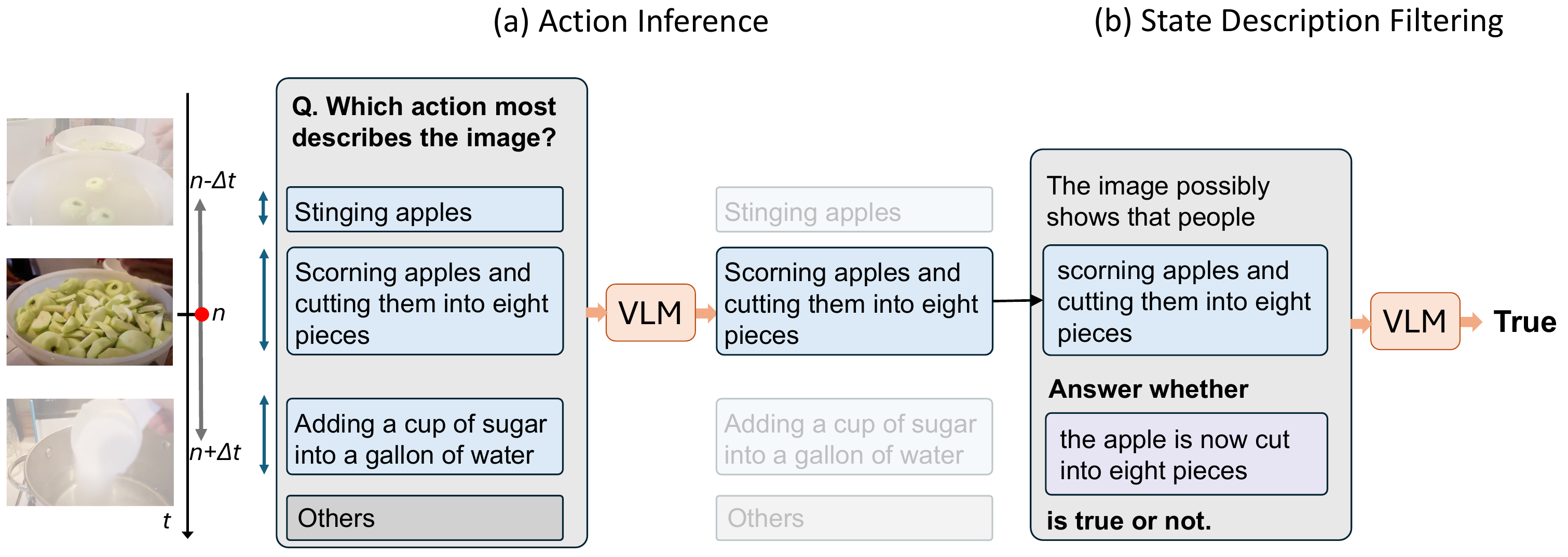}
    \caption{Temporal alignment between video frames and action/state description. (a) Given video frame, we select most likely actions from candidates using VLMs. (b) Interval is further filtered by matching frame and state description.}
    \label{fig:alignment}
\end{figure*}

First, given a video frame, we ask the VLM the most plausible manipulation action from candidate manipulation actions within before/after $\Delta t$ seconds around the frame (Figure~\ref{fig:alignment} (a)). We add ``others'' option for frames lacking action, which are then omitted from the training.

Once we assign likely actions to each frame, we then use the corresponding object state description to filter out irrelevant frames using VLMs (Figure~\ref{fig:alignment} (b)).

We use LLaVA-1.5 \cite{liu2023improved} as VLM, and each video frame is fed to LLaVA in its original resolution of 240p with text prompts to assign the correct action and the state description. We provide the prompts for action inference and state description filtering in Table \ref{tab:prompt_llava_action} and Table \ref{tab:prompt_llava_state}, respectively. For action inference, we prepare action candidates and prompt LLM to choose one of them. For state description filtering, the inferred action is added as context for a better understanding of the given video frame.

\clearpage
\section{Additional Information on Self-Training}
\label{supsec:self-training}
\begin{figure}[t]
    \centering
    \includegraphics[width=1\linewidth]{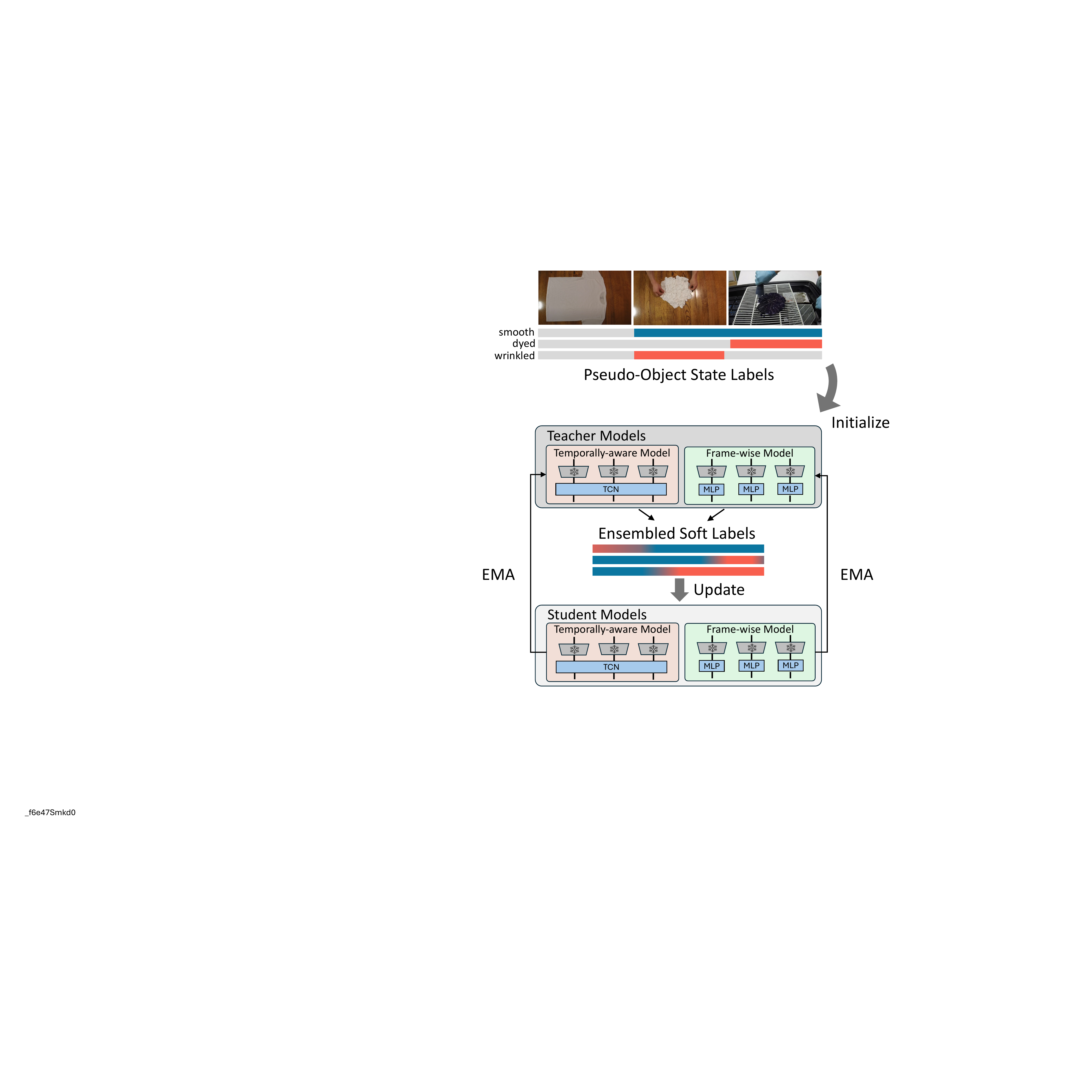}
    \caption{Self-training via ensemble of teacher models.} 
    \label{fig:self-training}
\end{figure}

Figure~\ref{fig:self-training} shows the overview of training.
We first train two frame-wise {\it teacher} state classification models with different temporal windows: {\bf MLP} and {\bf TCN}, using the original pseudo-labels.
While they are similar to action segmentation models (\eg, \cite{farha2019ms}), the final layer of each model has a sigmoid activation because object states are not mutually exclusive.
{\bf MLP} has a 2-layer MLP head and it does not take any temporal context into account.
{\bf TCN} has stacked temporal convolutional networks to model the temporal relationship between frames.
We modify MS-TCN~\cite{farha2019ms} to replace the softmax layer in each stage with a sigmoid layer.
At this phase, {\bf TCN} does not perform well due to missing pseudo-labels.

Given the above two models as teacher models, we propose to train another set of {\bf MLP} and {\bf TCN} as student models, taking the weighted sum of teacher models' predictions as ground truth.
Formally, the new ensembled target $\mathbf{y_t^{ens}}$ for $t$-th video frame can be denoted as follows:
\begin{align}
\mathbf{y_t^{ens}} = \mathbf{\alpha \cdot \hat{y}_t^{TCN}}+ (1-\alpha) \cdot \mathbf{\hat{y}_t^{MLP}}.
\end{align}
where $\mathbf{\hat{y}_t^{TCN}}$ and $\mathbf{\hat{y}_t^{MLP}}$ are the output of {\bf MLP} and {\bf TCN}, respectively, and $\alpha \in [0, 1]$ denote the weight of different labels.
In this work, we set $\alpha$ to 0.5.

Following the mean teacher strategy~\cite{tarvainen2017mean,han2022temporal}, we concurrently update the parameters of the teacher models via the Exponential Moving Average (EMA) mechanism.
The teacher models are gradually updated to follow the student models.
We use the momentum coefficient of 0.999.
After student model training, only the student {\bf TCN} model is used for inference.

\section{Additional Information on Experiments}
\label{supsec:experiments}
\subsection{Experimental Setup}
\subsubsection{Training Data Collection}

We used the subset of HowTo100M~\cite{howto100m} as training data.
We assume access to the target object names and their states to recognize.
Given a target object name (\eg, apple), we first selected videos that included the object name in the title and narrations.
We further selected videos that included the verbs likely to be associated with the target objects.
We used GPT-4~\cite{achiam2023gpt} for creating the verb list for subset selection.
Table~\ref{tab:prompt_list_verbs} shows the prompt for listing associated verbs.
The resulting verb list is shown in Table \ref{tab:associated_verbs}. 
Finally, we removed videos with very long narrations ($>$ 12,000 words) to fit the maximum context length of LLMs, resulting in around 2K videos per object category by adjusting the maximum YouTube search rank.

For pseudo-labeling, we used GPT-3.5 (\texttt{gpt-3.5-turbo-1106}) to process narrations. We provide the number of training videos, the hours of videos, and the pseudo-label assignment rate in Table \ref{tab:training_data}.


\begin{table}[t]
    \centering
    \begin{tcolorbox}
    
    \small{
    List as many verbs that describe actions associated with each object state. Associated actions include the actions that are necessary to produce the object state and actions commonly performed on objects in that state.\\

    \textcolor{blue}{\{list of object states\}}\\

    The answer format should be in comma-delimited CSV format. The verbs should be in infinitive form and single word.\\
Foramt: "$<$object state$>$(completely the same as given state description)","$<$verb$>$,$<$verb$>$,...,$<$verb$>$"
    }
    \end{tcolorbox}
    \caption{Prompt to list associated verbs. \textcolor{blue}{\{list of object states\}} with list of object state texts (\eg \textit{The egg is raw}) concatenated by newline.}
    \label{tab:prompt_list_verbs}
\end{table}
\begin{table}[t]
    \centering
    \begin{tabular}{lrrr}
    \toprule
        Category & \parbox[t]{4em}{\raggedleft  Number\\of videos} & \parbox[t]{4em}{\raggedleft  Video duration\\(hours)} & \parbox[t]{6em}{\raggedleft Label \\ assignment rate}\\
    \midrule
        Apple & 2095& 224 & 86\%\\
        Egg & 2379 & 229 & 64\%\\
        Flour & 1091 & 114 & 81\%\\
        Shirt & 1451 & 160 & 78\%\\
        Tire & 1520 & 167 & 67\%\\
        Wire & 1630 & 229 & 73\%\\
    \midrule
        Total & 10166 & 1123 & 74\%\\
    \bottomrule
    \end{tabular}
    \caption{Statistics of training data. Label assignment rate denotes the percentage of frames with either positive or negative labels assigned, averaged across all object state categories.}
    \label{tab:training_data}
\end{table}

\begin{table}[t]
    \centering
    \begin{tabular}{|l|p{1.8cm}|p{3.4cm}|}
    \hline
    Category & \parbox[t]{2cm}{  Primary\\Object Name} & Secondary Object Name\\
    \hline
        Apple & apple & apples, apple sauce, apple pie\\
    \hline
        Egg & egg & eggs, egg mixture, meringue, omelet, soufflée, cake, quish\\
    \hline
        Flour & flour & dough, pizza, pancake, fritter\\
    \hline
        Shirt & shirt & t-shirt, clothes\\
    \hline
        Tire & tire & tires, wheel \\
    \hline
        Wire & wire & wires, cable, ring, wire accessory\\
    \hline
    \end{tabular}
    \caption{Object names used for assigning background labels using CLIP model. In addition to the primary object name (second column), additional secondary object names are used for filtering (third column).}
    \label{tab:object_names}
\end{table}

\begin{table*}
\centering
    \begin{tabular}{|l|l|p{4cm}|}
    \hline
Object & State & Associated Verbs \\
\hline
\multirow{11}{*}{apple}
 & \small{whole} & \small{buy, pick, choose, handle, display, store} \\
\cline{2-3}
 & \small{peeled} & \small{peel, slice, cut, core, dip} \\
\cline{2-3}
 & \small{grated} & \small{grate, mix, stir, press, dry} \\
\cline{2-3}
 & \small{cut} & \small{cut, chop, slice, dice, core} \\
\cline{2-3}
 & \small{sliced} & \small{slice, serve, arrange, layer, present} \\
\cline{2-3}
 & \small{cooked} & \small{cook, boil, stew, simmer, heat} \\
\cline{2-3}
 & \small{baked} & \small{bake, heat, glaze, cover, fill} \\
\cline{2-3}
 & \small{smooth} & \small{blend, puree, whip, process, stir} \\
\cline{2-3}
 & \small{dried} & \small{dry, store, package, grind, rehydrate} \\
\cline{2-3}
 & \small{mixed ...} & \small{mix, stir, blend, incorporate, season} \\
\cline{2-3}
 & \small{caramelized} & \small{caramelize, heat, stir, glaze, serve} \\
\hline
\multirow{13}{*}{egg}
 & \small{intact} & \small{handle, observe, inspect} \\
\cline{2-3}
 & \small{raw} & \small{crack, beat, whisk} \\
\cline{2-3}
 & \small{boiled} & \small{peel, eat, cool} \\
\cline{2-3}
 & \small{poached} & \small{serve, eat, season} \\
\cline{2-3}
 & \small{cracked} & \small{break, separate, whisk} \\
\cline{2-3}
 & \small{runny} & \small{stir, cook, season} \\
\cline{2-3}
 & \small{whisked} & \small{pour, mix, beat} \\
\cline{2-3}
 & \small{cooked} & \small{eat, serve, cut} \\
\cline{2-3}
 & \small{fried} & \small{flip, serve, eat} \\
\cline{2-3}
 & \small{scrambled} & \small{stir, eat, serve} \\
\cline{2-3}
 & \small{foamy} & \small{beat, fold, stir} \\
\cline{2-3}
 & \small{baked} & \small{remove, eat, serve} \\
\cline{2-3}
 & \small{mixed ...} & \small{stir, whisk, bake} \\
\hline
\multirow{8}{*}{flour}
 & \small{dry} & \small{measure, sift, pour, store} \\
\cline{2-3}
 & \small{mixed ...} & \small{stir, whisk, blend, mix} \\
\cline{2-3}
 & \small{combined ...} & \small{combine, mix, beat, incorporate} \\
\cline{2-3}
 & \small{lumpy} & \small{sift, beat, whisk, crush} \\
\cline{2-3}
 & \small{kneaded} & \small{knead, punch, stretch, fold} \\
\cline{2-3}
 & \small{flattened} & \small{roll, press, flatten, spread} \\
\cline{2-3}
 & \small{baked} & \small{bake, cool, cut, serve} \\
\cline{2-3}
 & \small{fried} & \small{fry, turn, drain, serve} \\
\hline
\end{tabular}
\hfill
\begin{tabular}{|l|l|p{4cm}|}
\hline
Object & State & Associated Verbs \\
\hline
\multirow{10}{*}{shirt}
 & \small{stained} & \small{spill, drop, smear, splash, soak} \\
\cline{2-3}
 & \small{soaked} & \small{drench, immerse, submerge, squirt, splash} \\
\cline{2-3}
 & \small{dry} & \small{air-dry, blow-dry, heat, wring, squeeze} \\
\cline{2-3}
 & \small{wrinkled} & \small{crumple, crush, squeeze, wear, curl} \\
\cline{2-3}
 & \small{smooth} & \small{iron, press, flatten, stretch, pull} \\
\cline{2-3}
 & \small{dyed} & \small{color, soak, immerse, dip, stain} \\
\cline{2-3}
 & \small{folded} & \small{crease, bend, pleat, wrap, tuck} \\
\cline{2-3}
 & \small{piled} & \small{stack, heap, accumulate, gather, assemble} \\
\cline{2-3}
 & \small{rolled} & \small{spin, rotate, turn, scroll, wrap} \\
\cline{2-3}
 & \small{hung} & \small{suspend, drape, fasten, hook, display} \\
\hline
\multirow{7}{*}{tire}
 & \small{flat} & \small{inflate, inspect, replace, repair} \\
\cline{2-3}
 & \small{rinsed} & \small{spray, scrub, wash, dry} \\
\cline{2-3}
 & \small{taken off ...} & \small{unscrew, remove, lift, lower} \\
\cline{2-3}
 & \small{nuts ...} & \small{unscrew, replace, remove, loose} \\
\cline{2-3}
 & \small{covered ...} & \small{spray, spread, rub, clean} \\
\cline{2-3}
 & \small{dirty} & \small{wash, scrub, rinse, clean} \\
\cline{2-3}
 & \small{mounted ...} & \small{fit, secure, align, install} \\
\hline
\multirow{10}{*}{wire}
 & \small{straight} & \small{straighten, align, smooth, flatten} \\
\cline{2-3}
 & \small{twisted} & \small{twist, coil, bend, entangle} \\
\cline{2-3}
 & \small{soldered} & \small{solder, join, attach, adhere} \\
\cline{2-3}
 & \small{cut} & \small{cut, truncate, sever, divide} \\
\cline{2-3}
 & \small{damaged} & \small{damage, hit, break, puncture} \\
\cline{2-3}
 & \small{exposed} & \small{expose, uncover, reveal, unveil} \\
\cline{2-3}
 & \small{coiled} & \small{coil, roll, loop, curl} \\
\cline{2-3}
 & \small{braided} & \small{braid, twist, weave, interlace} \\
\cline{2-3}
 & \small{looped} & \small{loop, circle, bend, round} \\
\cline{2-3}
 & \small{connected} & \small{connect, link, tie, join} \\
\hline
\end{tabular}
\caption{List of associated verbs for each object state. This information is used to find videos to use in training.}
\label{tab:associated_verbs}
\end{table*}

\subsubsection{Details on Baseline Models}
For the LLaVA baseline, we fed each video frame and text prompt in Table \ref{tab:prompt_llava_baseline} to the model. The output from LLaVA is parsed and we use only True or False for prediction.
For the CLIP baseline, we input each video frame and object state text to the image and text encoder to calculate the similarity score between frames and object states, and use it as a confidence score. 
We tested several text prompts: ``a photo of \textless object state\textgreater \textless primary object name\textgreater'', ``The \textless primary object name\textgreater is \textless object state\textgreater'', ``a photo of \textless primary/secondary object name\textgreater'', and ``\textless object state description\textgreater''. The best prompt is chosen for the object category, which maximizes the average score in the category.
For the InternVideo baseline, we input the model in 8~fps and the object state text in the same way as CLIP.
\begin{table}[t]
    \centering
    \begin{tcolorbox}
    \small{
    Answer whether the image shows that "\textcolor{blue}{\{state\}}". First describe the image. Then, specify the judgement with "The answer is True/False."\\

    \textcolor{blue}{\{description\}}
    }
    \end{tcolorbox}
    \caption{Prompt for zero-shot object state inference using LLaVA. We replace \textcolor{blue}{\{state\}} with object state text (\eg \textit{The egg is raw}), \textcolor{blue}{\{description\}} with description of state.}
    \label{tab:prompt_llava_baseline}
\end{table}

\subsubsection{Implementation Details}
When generating pseudo-labels, we use CLIP~\cite{clip} to assign negative labels to the frames without target objects by taking the cosine similarity between a video frame and pre-defined target object names. Since the target object changes its states and can be called in another way, we manually set several object names for each category as shown in Table \ref{tab:object_names}. The object names are then used in text prompts as ``a photo of \textless object name\textgreater''. We take the maximum similarity score across the different prompts for each frame and compare it to the threshold of 0.2.

Our proposed \textbf{TCN} model consists of the pre-trained InternVideo visual encoder (InternVideo-MM-14-L) \cite{wang2022internvideo} followed by a modified MS-TCN \cite{farha2019ms} classifier. MS-TCN is a multi-stage temporal convolutional network where each stage outputs a prediction that is refined by the subsequent stages. We implement 4 stages where each stage consists of 10 layers of 512 dimensions followed by a final linear layer of $K$ outputs with sigmoid activation. Dropout value of 0.5 is used after each layer. We replace the softmax activation with sigmoid to match the multi-label setting.


We have two learning stages: one for learning the teacher models using pseudo-labels generated by LLMs and the other for self-training with the student model. We train each model with a batch size of 16 for 50 epochs in the first stage, followed by 50 epochs in the second one. We use the Adam optimizer~\cite{kingma2014adam} with decoupled weight decay regularization~\cite{loshchilov2017decoupled}. We set the learning rate to 1e-4 and the weight decay to 0.01.

\subsection{Results}
\subsubsection{Qualitative Results of Pseudo-Labeling via LLMs}
\label{subsubsec:llm_ex}
\begin{figure*}[t]
    \centering
    \includegraphics[width=0.9\textwidth]{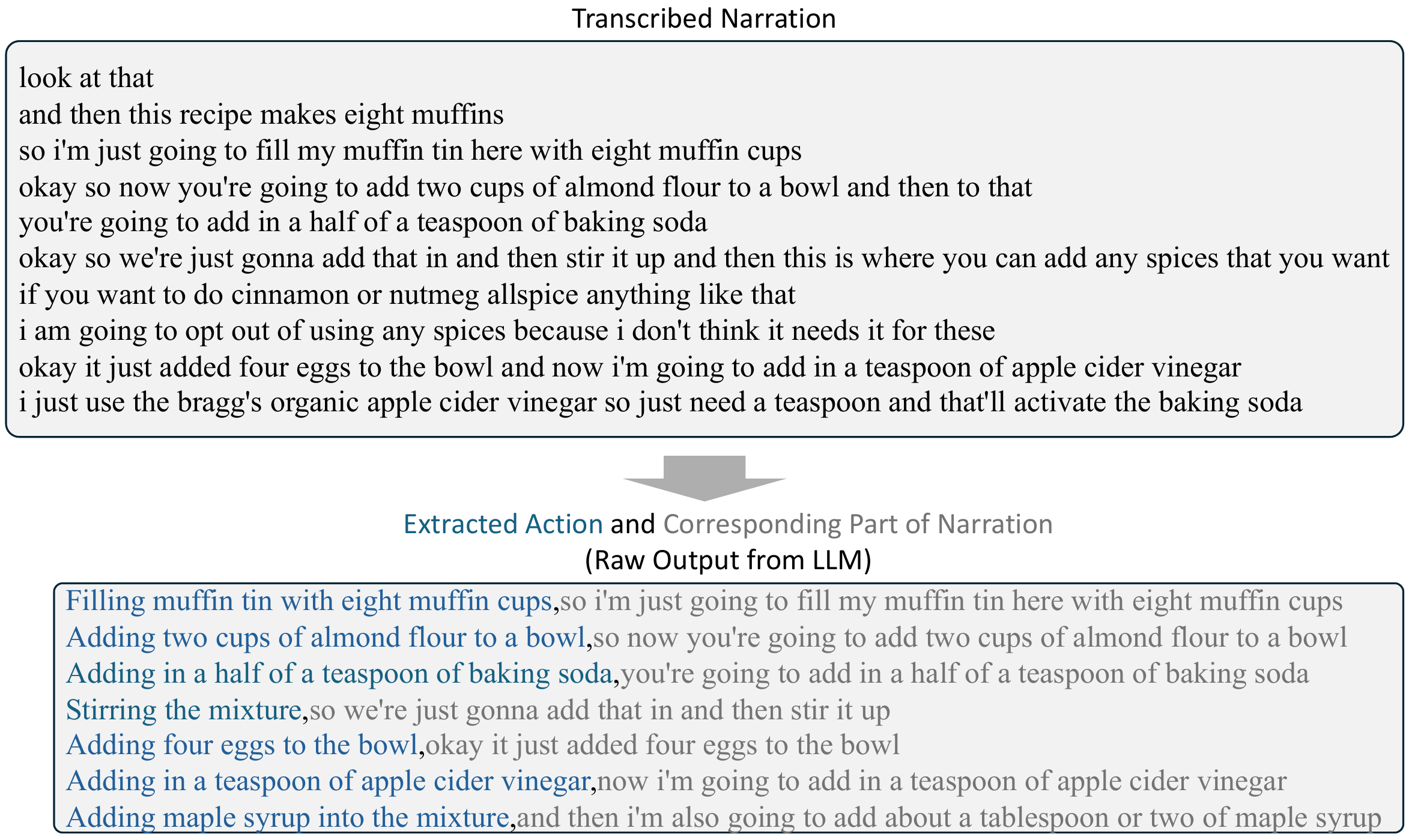}
    \caption{Examples of actual input narrations and LLM outputs for action.}
    \label{fig:llm_ex_action}
\end{figure*}

\begin{figure*}[t]
    \centering
    \includegraphics[width=0.9\textwidth]{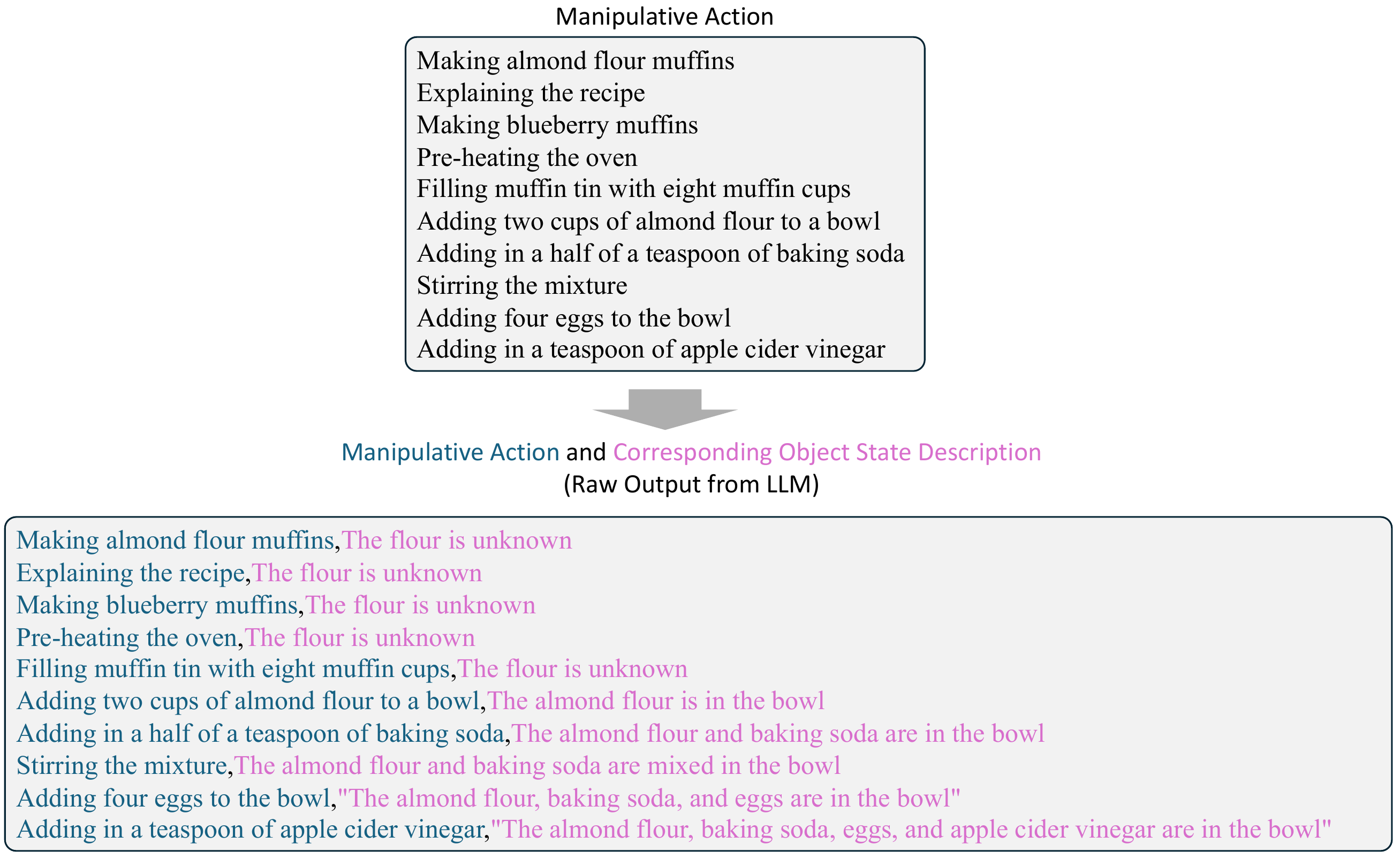}
    \caption{Examples of actual manipulative action and LLM outputs for object state description.}
    \label{fig:llm_ex_sd}
\end{figure*}

\begin{figure*}[t]
    \centering
    \includegraphics[width=0.9\textwidth]{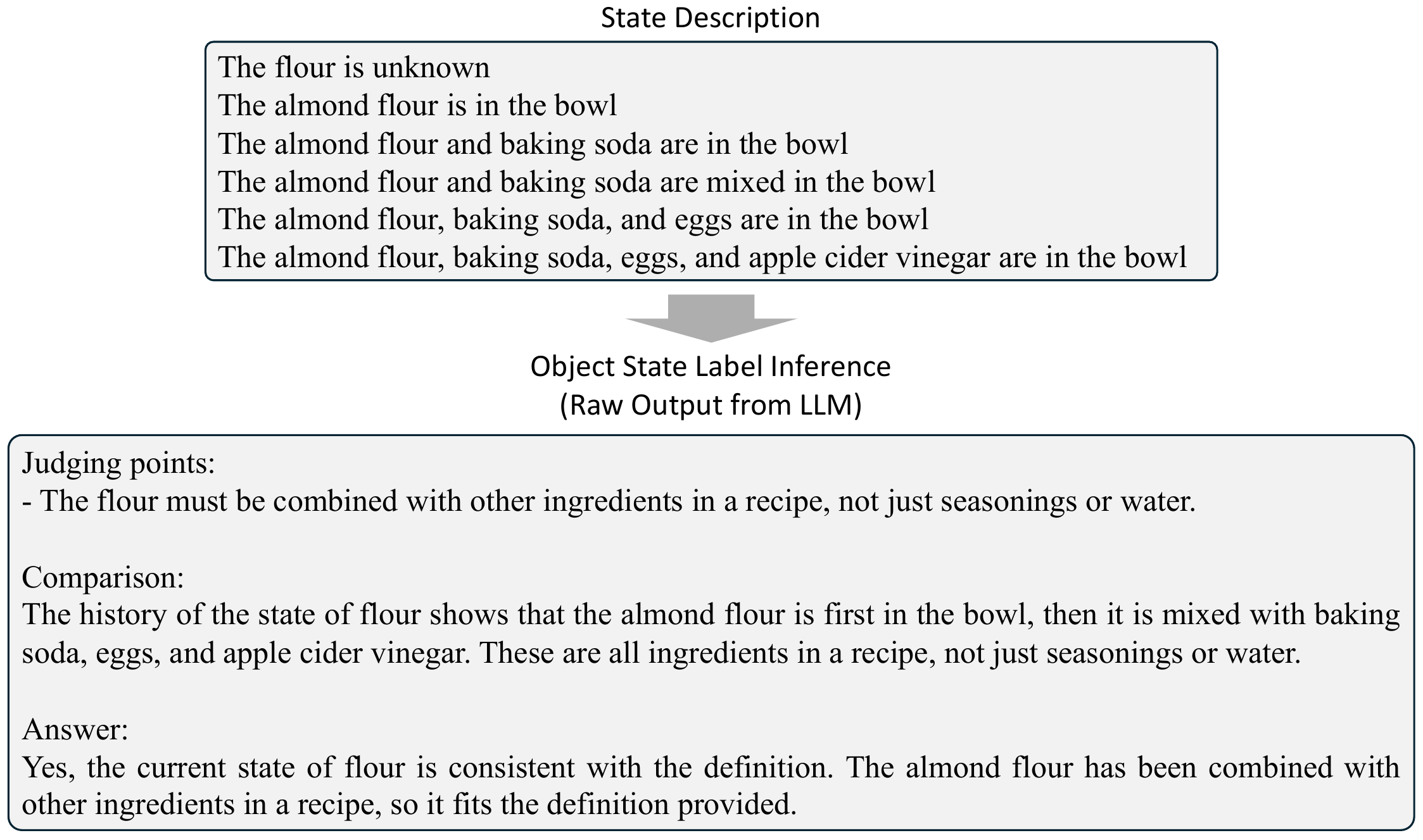}
    \caption{Examples of actual state description and LLM outputs for object state label inference.}
    \label{fig:llm_ex_label}
\end{figure*}

Figure \ref{fig:llm_ex_action}, \ref{fig:llm_ex_sd}, and \ref{fig:llm_ex_label} provide the raw output from LLMs in the process of (a) extracting manipulation actions from narrations, (b) generating state descriptions from the actions, and (c) inferring context-aware object state labels, respectively.

\subsubsection{Additional Qualitative Results}
\begin{figure*}[t]
    \centering
    \includegraphics[width=1\linewidth]{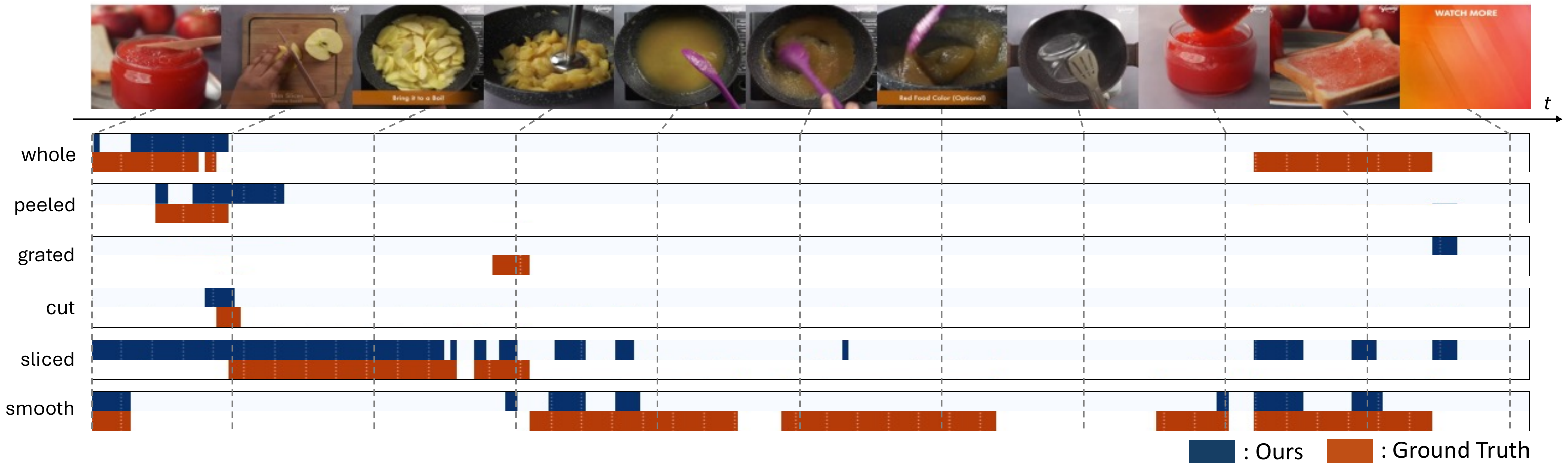}
    \caption{Qualitative results of proposed model (target object: apple).}
    \label{fig:fq_apple}
\end{figure*}

\begin{figure*}[t]
    \centering
    \includegraphics[width=1\linewidth]{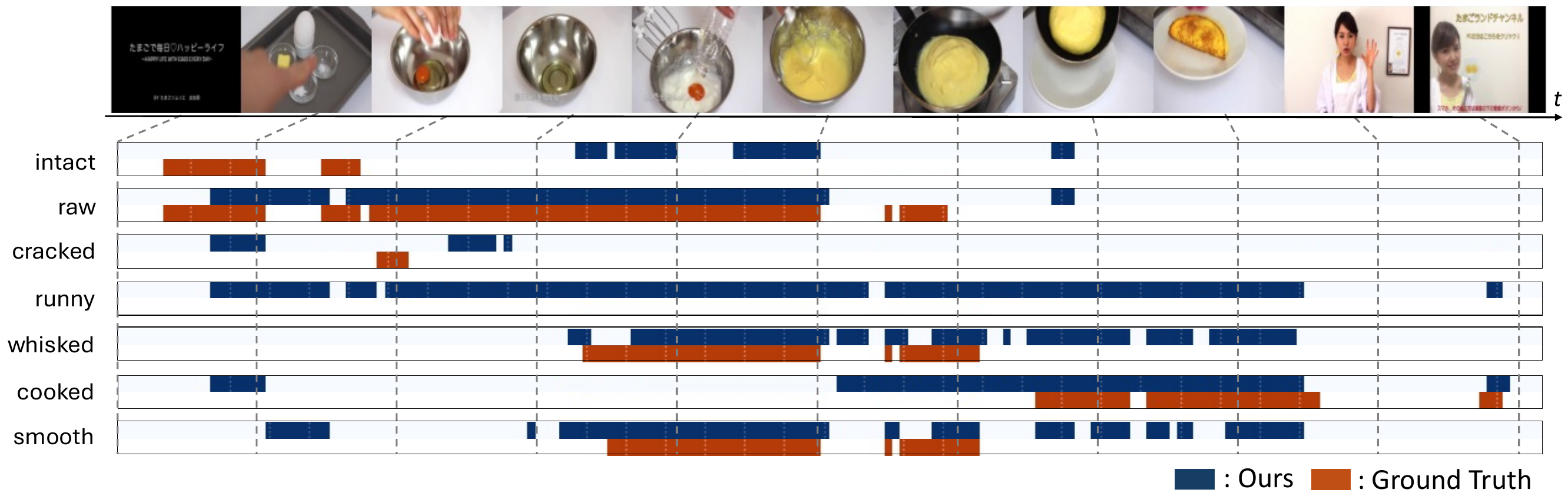}
    \caption{Qualitative results of proposed model (target object: egg).}
    \label{fig:fq_egg}
\end{figure*}

\begin{figure*}[t]
    \centering
    \includegraphics[width=1\linewidth]{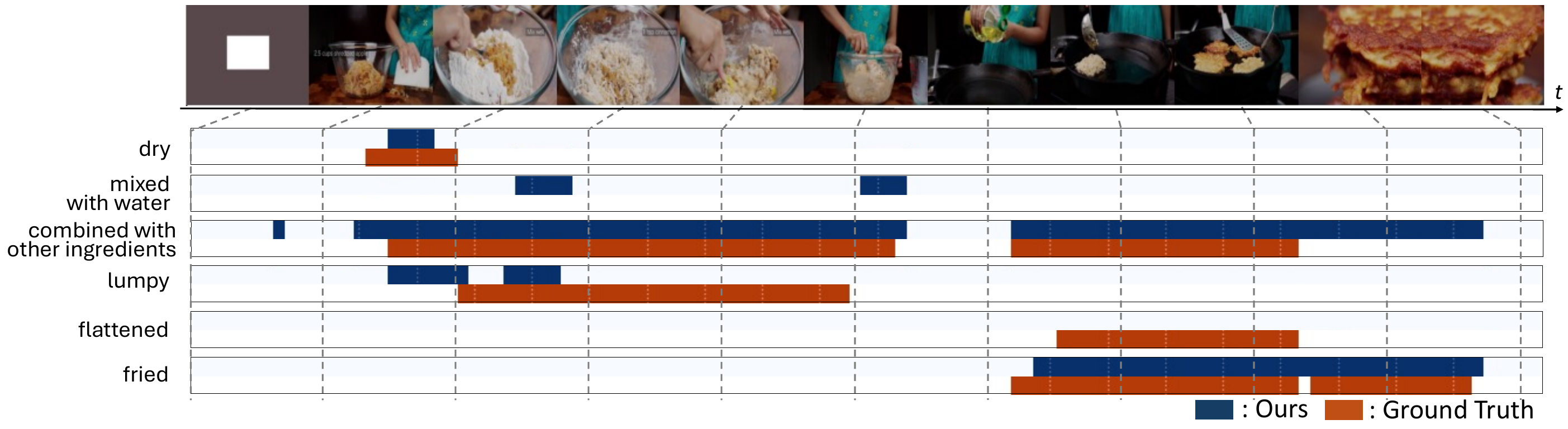}
    \caption{Qualitative results of proposed model (target object: flour).}
    \label{fig:fq_flour}
\end{figure*}

\begin{figure*}[t]
    \centering
    \includegraphics[width=1\linewidth]{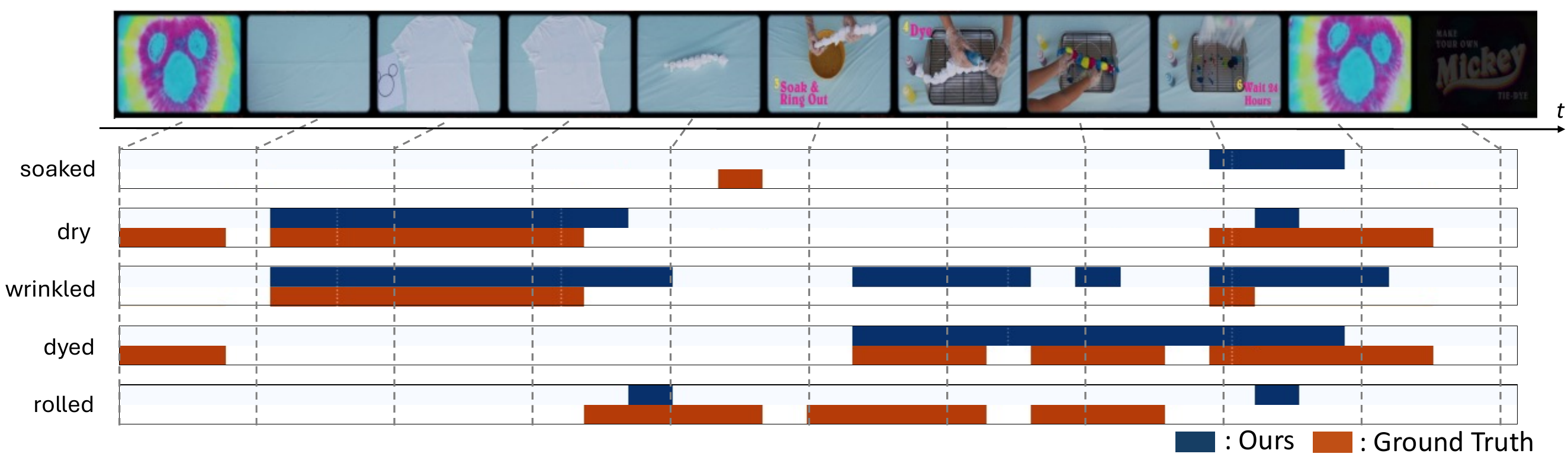}
    \caption{Qualitative results of proposed model (target object: shirt).}
    \label{fig:fq_shirt}
\end{figure*}

\begin{figure*}[t]
    \centering
    \includegraphics[width=1\linewidth]{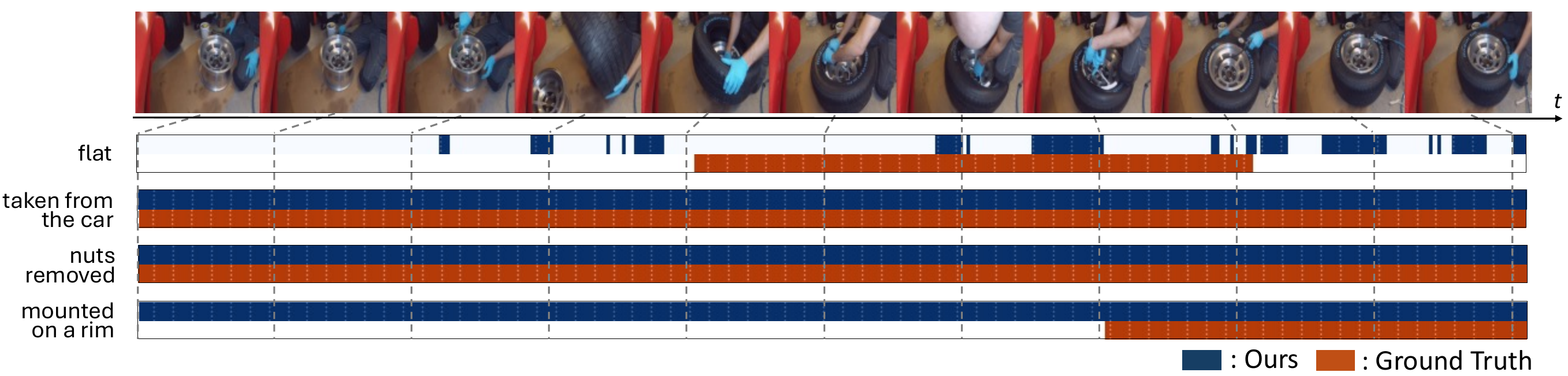}
    \caption{Qualitative results of proposed model (target object: tire).}
    \label{fig:fq_tire}
\end{figure*}

\begin{figure*}[t]
    \centering
    \includegraphics[width=1\linewidth]{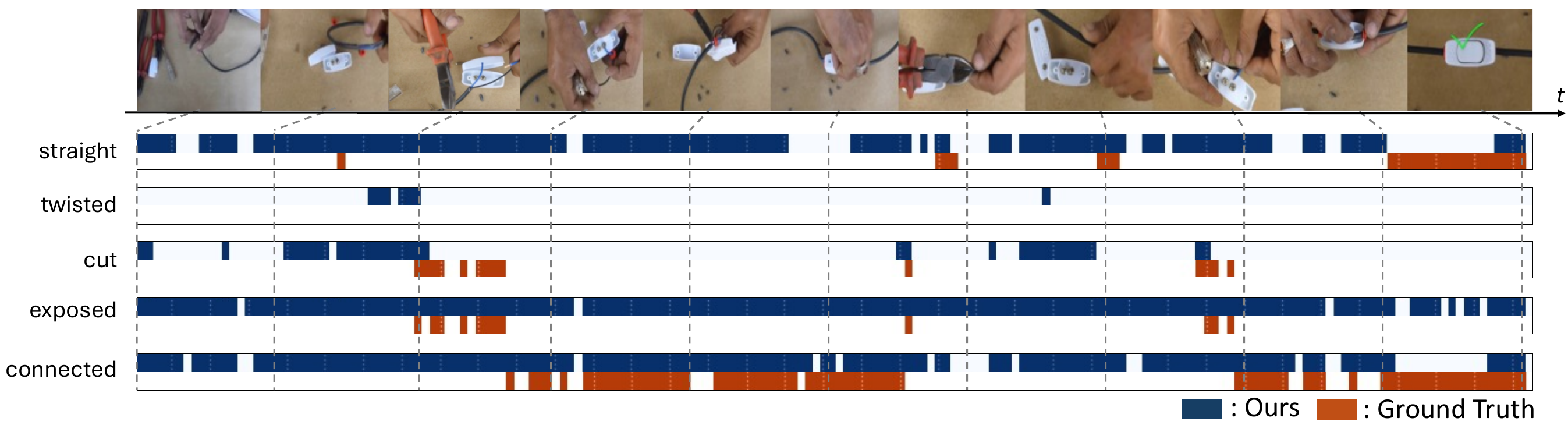}
    \caption{Qualitative results of proposed model (target object: wire).}
    \label{fig:fq_wire}
\end{figure*}

We provide qualitative results in comparison with our model's predictions and the ground truth in Figure \ref{fig:fq_apple} - \ref{fig:fq_wire}. We pick up a particularly relevant state in each video and display the results. For all tables, blue bars represent predictions, and red bars represent ground truth. 

\subsubsection{Failure Case in Pseudo-Labeling}
\begin{figure*}[tb]
    \centering
    \includegraphics[width=0.9\linewidth]{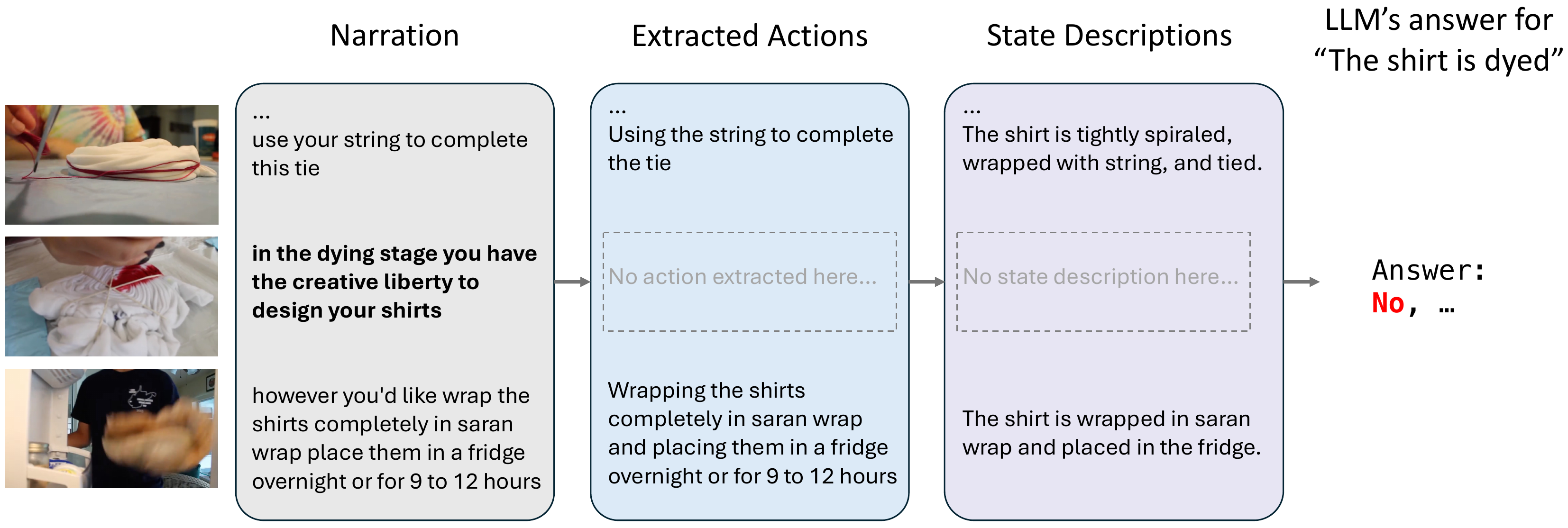}
    \caption{Example of failure case where key action is overlooked due to less specific description in narration, leading to false negatives for object state ``The shirt is dyed". The second sentence in narration could indicate that the action of “dying” is actually occurring in the video, or it might simply be explaining the creative liberty in dying process. Our framework sometimes interprets explanations as unrelated to actions happening in video.}
    \label{fig:failure_case}
\end{figure*}

We provide a typical failure case in Figure \ref{fig:failure_case}. LLMs sometimes fail to extract key actions due to less specific descriptions in narration, leading to false negatives for object states.


\clearpage
\subsubsection{Per-state Results}


For an in-depth understanding of the proposed method, we report per-state performance in Table~\ref{tab:result_taxonomy}.
We categorize the object-state pairs into four groups in terms of their position in the task of instructional videos.

\begin{enumerate}[label=\alph*),ref=\alph*]
    \item \textbf{Typical goal state.} The object states that are commonly present as the final state of certain tasks, such as \textit{baked apple/egg/flour}, \textit{scrambled egg}, \textit{fried egg/flour}, \textit{tire mounted on a rim}, and \textit{soldered wire}. We observe strong performance for this type. One reason is that these states are crucial in instructional videos, often containing the necessary action information to achieve these states in the narrations. Additionally, these states almost do not change due to subsequent actions, eliminating the need for complex reasoning.
    \item \textbf{Intermediate state.} The intermediate object states in certain tasks, such as \textit{peeled apple}, \textit{cut apple}, \textit{cracked egg}, \textit{rinsed tire}, and \textit{twisted wire}. We observe moderate performance for this type. Since people sometimes skip mentioning intermediate actions included in a task in a video, it is sometimes more difficult to assign correct state labels than the goal states. Furthermore, the presence of states may change several times due to subsequent actions, which makes it complicated to assign labels.
    \item \textbf{Typical initial state.} The object state that would normally exist initially if nothing were done to the object, such as \textit{whole apple,} \textit{intact egg}, \textit{raw egg}, and \textit{dry flour}. We surprisingly observe strong performance for this type. We initially thought that it is difficult for LLMs to infer these states because they don't emerge as a result of a certain action. However, LLMs can successfully infer these classes from actions that are not directly related. For example, ``picking apples from the tree'' implies that the apple is in whole state.
    \item \textbf{Unintended state in instructional videos.} The object states that are not usually intended in instructional videos, such as \textit{damaged wire}, \textit{dirty tire}, and \textit{stained shirt} (\ie there are few instructional videos that intentionally damage wire). We observe poor performance for this type.
\end{enumerate}

\begin{table}
    \scalebox{0.76}{
    \begin{tabular}{lrr}
    \toprule
    \multirow{3}{*}{Object State} & \multicolumn{2}{c}{F1-max} \\
    \cmidrule(lr){2-3}
    & Ours & \hspace{1em}InternVideo\\
    \midrule
    \multicolumn{3}{l}{\textbf{(a.)} \textit{Typical goal state}} \\
     The apple is cooked. & \textbf{0.76} & 0.60\\
     The apple is baked. & \textbf{0.67} & 0.45\\
     The apple is caramelized. & \textbf{0.70} & 0.57\\
     The apple is dried. & 0.70 & 0.70\\
     The egg is boiled. & \textbf{0.64} & 0.44\\
     The egg is poached. & 0.52 & \textbf{0.59}\\
     The egg is cooked. & \textbf{0.77} & 0.71\\
     The egg is fried. & \textbf{0.64} & 0.35\\
     The egg is scrambled. & 0.63 & \textbf{0.85}\\
     The egg is baked. & \textbf{0.88} & 0.36\\
     The flour is baked. & \textbf{0.60} & 0.26\\
     The flour is fried. & \textbf{0.61} & 0.58\\
     The shirt is smooth. & \textbf{0.66} & 0.62\\
     The shirt is dyed. & \textbf{0.61} & 0.33\\
     The shirt is folded. & \textbf{0.63} & 0.60\\
     The shirt is piled. & \textbf{0.39} & 0.17\\
     The shirt is hung. & 0.32 & \textbf{0.38}\\
     The tire is flat. & \textbf{0.59} & 0.51\\
     The tire is taken from the car. & \textbf{0.89} & 0.87\\
     The tire is mounted on a rim. & 0.90 & 0.90\\
     The wire is soldered. & \textbf{0.65} & 0.52\\
     The wire is braided. & \textbf{0.61} & 0.27\\
     The wire is looped. & \textbf{0.81} & 0.30\\
    \midrule
    \multicolumn{3}{l}{\textbf{(b.)} \textit{Intermediate state}} \\
     The apple is peeled. & 0.32 & \textbf{0.35}\\
     The apple is grated. & 0.12 & \textbf{0.26}\\
     The apple is cut. & 0.56 & \textbf{0.68}\\
     The apple is sliced. & \textbf{0.76} & 0.74 \\
     The apple is smooth. & \textbf{0.45} & 0.17\\
     The apple is mixed with other ingredients. & \textbf{0.18} & 0.15\\
     The egg is cracked.  & 0.19 & \textbf{0.26}\\
     The egg is runny. & 0.23 & \textbf{0.25}\\
     The egg is whisked. & 0.52 & \textbf{0.60}\\
     The egg is foamy. & \textbf{0.48} & 0.19\\
     The egg is mixed with other ingredients. & \textbf{0.40} & 0.31\\
     The flour is mixed with water. & 0.41 & \textbf{0.60}\\
     The flour is combined with other ingredients. & 0.31 & 0.31\\
     The flour is lumpy. & \textbf{0.62} & 0.58\\
     The flour is kneaded. & \textbf{0.52} & 0.38\\
     The flour is flattened. & \textbf{0.67} & 0.34\\
     The shirt is soaked & \textbf{0.52} & 0.50\\
     The shirt is wrinkled & 0.32 & \textbf{0.36}\\
     The shirt is rolled. & \textbf{0.45} & 0.22\\
     The tire is rinsed.  & \textbf{0.43} & 0.34\\
     The tire is nuts removed. & \textbf{0.80} & 0.79\\
     The tire is covered with wheel cleaner. & \textbf{0.48} & 0.37\\
     The wire is twisted. & 0.47 & \textbf{0.54}\\
     The wire is cut. & 0.30 & \textbf{0.35}\\
     The wire is exposed. & 0.69 & \textbf{0.82}\\
     The wire is coiled. & 0.49 & \textbf{0.50}\\
     The wire is connected. & 0.19 & \textbf{0.26}\\
    \midrule
    \multicolumn{3}{l}{\textbf{(c.)} \textit{Typical initial state}} \\
     The apple is whole. & \textbf{0.59} & 0.40\\
     The egg is raw.  & \textbf{0.72} & 0.48\\
     The egg is intact. & 0.27 & \textbf{0.34}\\
     The flour is dry.  & \textbf{0.66} & 0.42\\
    \midrule
    \multicolumn{3}{l}{\textbf{(d.)} \textit{Unintended state in instructional videos}} \\
     The shirt is stained.  & \textbf{0.40} & 0.15\\
     The tire is dirty.  & \textbf{0.21} & 0.20\\
     The wire is straight. & \textbf{0.43} & 0.29\\
     The wire is damaged. & \textbf{0.23} & 0.11\\
    \bottomrule
    \end{tabular}
    }
\caption{Per-state category results. Each object state category is assigned to one of four subgroups.}
\label{tab:result_taxonomy}
\end{table}

\subsection{Experiments on ChangeIt Dataset}
\label{supsubsec:changeit}
ChangeIt dataset \cite{look_for_the_change} evaluates the performance of temporally localizing (i) the initial state, (ii) the state-modifying action, and (iii) the end state.
For 44 state-changing actions, 667 videos up to 48 hours are annotated for evaluation. 

We slightly modify our method to fit the label space of ChangeIt dataset. Specifically, we change the prompts for state label inference as shown in Table \ref{tab:prompt_changeit}. Since ChangeIt dataset only includes the state-changing action category names, we prepared several adjectives that represent the end state of each action category. In addition, for \textit{background} label, we compute the threshold between 0.16 to 0.25 and report the best results since the annotation rules of the \textit{background} label are not unified, and are different for each object category in the dataset. Due to high computational demands, we omit the matching between state description and video frames.
For a fair comparison, we use the causal ordering constraints~\cite{look_for_the_change} only at inference by selecting the frames of \textit{initial state}, \textit{action}, and \textit{end state} with their highest probabilities while respecting their causal order.

For training dataset, we used the subset of the original ChangeIt videos including narrations for training. We selected at most 350 videos per category with the highest noise adaptive weight~\cite{look_for_the_change}.
This results in 10,749 videos in total, which is 32\% of the original dataset. 

For evaluation, we report precision@1 for action and state respectively following \cite{look_for_the_change}. For the results of baseline models, we use the scores reported in the original papers for the results of compared models in the ChangeIt dataset.

\begin{table}[t]
    \centering
    \begin{tcolorbox}
    \small{
    This is a history of state of \textcolor{blue}{\{object\}}:\\
\textcolor{blue}{\{list of state description up to that point\}}\\

You need to infer the state of the \textcolor{blue}{\{object\}} based on the history. When you answer, choose from the options below.\\
Options:\\
Initial - The \textcolor{blue}{\{object\}} is just before being \textcolor{blue}{\{end state\}}, but \textcolor{blue}{\{action\}} has not started.\\
Action - The \textcolor{blue}{\{object\}} is now being \textcolor{blue}{\{end state\}}.\\
End - The \textcolor{blue}{\{object\}} has already been \textcolor{blue}{\{end state\}}, and \textcolor{blue}{\{action\}} has been completed.\\
Ambiguous - Cannot identify the state from the action information, or the action is totally unrelated to \textcolor{blue}{\{action\}}.\\

Think step-by-step as follows:\\
- First, describe the current state of the object in detail based on the history.\\
- Then, answer Initial/Action/End/Ambiguous and reason.\\

Current State: [detailed state description]\\

Answer: [yes/no/ambiguous and why]
    }
    \end{tcolorbox}
    \caption{Prompt to infer object state label from descriptions for ChangeIt dataset. We replace \textcolor{blue}{\{object\}} with primary object name (Table \ref{tab:object_names}), \textcolor{blue}{\{list of state description up to that point\}} with state descriptions concatenated by newline, \textcolor{blue}{\{end state\}} with prepared end states concatenated by comma, and \textcolor{blue}{\{action\}} with state-changing action category name. Black text is used as is.}
    \label{tab:prompt_changeit}
\end{table}

\end{document}